\documentclass[ai,article,accept,pdftex,moreauthors,ai]{Definitions/mdpi}
\firstpage{1}
\pubvolume{6}
\issuenum{4}
\articlenumber{69}
\pubyear{2025}
\copyrightyear{2025}
\externaleditor{ Xianrong (Shawn) Zheng}
\datereceived{17 February 2025}
\daterevised{20 March 2025} % Comment out if no revised date
\dateaccepted{28 March 2025}
\datepublished{3 April 2025}
\hreflink{https://doi.org/10.3390/ai6040069} % If needed use \linebreak
\usepackage{algorithm}  % PCH
\usepackage{algorithmic}  % PCH
\usepackage{amsfonts}   % PCH
\usepackage{amsmath}    % PCH
\usepackage{multirow}    % PCH
\usepackage{xcolor}    % PCH
\usepackage{float}    % PCH
\usepackage{wrapfig}    % PCH
\usepackage{graphicx}    % PCH
\usepackage{subcaption}    % PCH
\usepackage{pgfplots}    % PCH
\usepackage{fontawesome}    % PCH

\pgfplotsset{width=0.55\textwidth, compat=1.18}   % PCH
\definecolor{mypink}{RGB}{255, 105, 180}    % PCH
\definecolor{myblue}{RGB}{50, 20, 255}    % PCH ; changes addressing reviewers' comments
\definecolor{mygreen}{RGB}{0, 128, 0}    % PCH ; changes addressing LSCM's comments
\definecolor{myred}{RGB}{255, 25, 50}    % PCH ; for my own notes

\definecolor{plotred}{RGB}{255, 0, 0}       % PCH
\definecolor{plotpink}{RGB}{255, 192, 203}  % PCH
\definecolor{plotblue}{RGB}{31, 119, 180}    % PCH
\definecolor{plotgreen}{RGB}{44, 160, 44}    % PCH
\definecolor{plotorange}{RGB}{255, 127, 14}    % PCH
\definecolor{plotpurple}{RGB}{148, 103, 189}    % PCH

\Title{LineMVGNN: Anti-Money Laundering with Line-Graph-Assisted Multi-View Graph Neural Networks}

\TitleCitation{LineMVGNN: Anti-Money Laundering with Line-Graph-Assisted Multi-View Graph Neural Networks}

\Author{{{Chung-Hoo Poon}} 
$^{1,2,}$*\href{https://orcid.org/0009-0001-7280-5911}{\orcidicon}, James Kwok $^{2}$\href{https://orcid.org/0000-0002-4828-8248}{\orcidicon}, Calvin Chow $^{1,}$* and Jang-Hyeon Choi $^{1}$}

\AuthorNames{Chung-Hoo Poon, James Kwok, Calvin Chow and Jang-Hyeon Choi}

\AuthorCitation{{{Poon, C.-H.;}} %MDPI: Please check all author names carefully.
Kwok, J.; Chow, C.; Choi, J.-H.}

\address{%
$^{1}$ \quad Logistics and Supply Chain MultiTech R\&D Centre, Level 11, Cyberport 2, 100 Cyberport Road, Hong Kong
\\
$^{2}$ \quad  {Department of Computer Science and Engineering, }The Hong Kong  {University} %MDPI: For universities, the department/school/faculty/campus is required. Please try to provide this information.
of Science and Technology, Clear Water Bay, Kowloon, Hong Kong
}

\corres{
Correspondence:
chpoonag@connect.ust.hk (C.-H.P.);
cchow@lscm.hk (C.C.)
}

\abstract{{{Anti-money}} %MDPI: This journals' Abstract should be structured to contain the subheadings. Please add.
laundering (AML) systems are important for protecting the global economy. However, conventional rule-based methods rely on domain knowledge, leading to suboptimal accuracy and a lack of scalability. Graph neural networks (GNNs) for digraphs (directed graphs) can be applied to transaction graphs and capture suspicious transactions or accounts. However, most spectral GNNs do not naturally support multi-dimensional edge features, lack interpretability due to edge modifications, and have limited scalability owing to their spectral nature. Conversely, most spatial methods may not capture the money flow well. Therefore, in this work, we propose LineMVGNN {(\underline{{Line}%MDPI: Please confirm if the underline should be retained. The highlights below are the same.
}-Graph-Assisted \underline{{M}}ulti-\underline{{V}}iew \underline{{G}}raph \underline{{N}}eural \underline{{N}}etwork)}, a novel spatial method that considers payment and receipt transactions.
Specifically, the LineMVGNN model extends a lightweight MVGNN module, which performs {two-way message passing} between nodes in a transaction graph. Additionally, LineMVGNN incorporates a {line graph view} of the original transaction graph to enhance the propagation of transaction information.
We conduct experiments on two real-world account-based transaction datasets: the {Ethereum phishing transaction network dataset} and a {financial payment transaction dataset} from one of our industry partners. The {results} show that our proposed method outperforms state-of-the-art methods, reflecting the effectiveness of money laundering detection with line-graph-assisted multi-view graph learning. We also discuss {scalability, adversarial robustness, and regulatory considerations} of our proposed method.
}


\begin{document}

%%%%%%%%%%%%%%%%%%%%%%%%%%%%%%%%%%%%%%%%%%
\section{Introduction}

Money laundering is the process of disguising the origins of illegally obtained funds to make them appear legitimate.
% round 1: reviewer 2: The introduction focuses on the problem of money laundering, highlighting how Graph Neural Networks (GNNs) have emerged as the de facto tool for fraud detection. A deeper investigation on the problem of money laundering detection could be appreciated.""
{Existing anti-money laundering (AML) systems can be generally categorized into two groups: rule-based methods and machine learning-based methods. Rule-based methods are popular among commercial institutions due to simple and easy-to-code rules predefined by domain experts \cite{10.1007/s10115-017-1144-z}.
However, it is time-consuming and labor-intensive to keep updating the rules under ever-evolving data.}
Machine learning-based anti-money laundering (AML) systems leverage large volumes of historical transaction data to learn models.
For instance, support vector machine, logistic regression, k-means clustering, k-nearest neighbors, random forests, MLP, etc. \cite{HILAL2022116429}.
In recent years, graph neural networks (GNNs) have emerged as the de facto tool for graph learning tasks.
Example application domains of GNNs for fraud detection tasks include credit card transactions, e-payment data, and cryptocurrency transactions \cite{MOTIE2024122156}.

% Example application domains of GNNs for fraud detection tasks include credit card transactions \cite{9204584,9540093,10.1007/978-3-030-37720-5_3,9436573}, e-payments data \cite{conf/ijcai/GongWSLN0P23,10.1145/3448016.3457564}, cryptocurrency transactions such as Ethereum transactions \cite{conf/wasa/DuanYDZY22,kanezashi2022ethereum,conf/www/LiGLHLX22,10.1016/j.future.2021.08.023,9896899,9667674,10.1007/978-3-030-65745-1_8} and Bitcoin transactions \cite{10016137,conf/blocksys/Tian0CSZ21,10.1145/3409073.3409080,Weber2019AntiMoneyLI}.

Some GNNs for digraphs (directed graphs) are suitable for money laundering detection
tasks since transaction graphs are directed. However, some not naturally supporting edge features are unsuitable for node-and-edge-attributed transaction graphs. This type of graph is common for account-based transaction graphs, in which nodes and edges represent accounts and transactions, respectively.

Recent state-of-the-art (SOTA) GNNs for digraphs can be categorized into {{{spectral methods}}%MDPI: Please confirm if the italics should be retained. The highlights below are the same.
} \cite{journals/corr/abs-2310-02232,10.1609/aaai.v37i6.25919,conf/nips/ZhangHBPH21,conf/nips/TongLSLR020,journals/corr/abs-2004-13970,Ma2019SpectralbasedGC,DBLP:journals/corr/abs-1802-01572} and {{{spatial
methods}}} \cite{conf/log/RossiCGFGB23,conf/iclr/Thost021,journals/corr/LiTBZ15}. Spectral GNNs perform graph convolutions by applying graph signal filters in the spectral domain, while spatial GNNs aggregate feature information from neighboring nodes directly in the spatial domain.

Although most of the recent works focus on the spectral methods, spectral GNNs for digraphs may not be suitable for attributed account-based transaction graphs,
because (1)~they do not naturally support multi-dimensional edge features, and (2) they have limited scalability due to their spectral nature. Usually, full graph propagation is needed during training, and the number of required graph propagations depends on the degree of the polynomial graph filter.

In contrast, spatial GNNs for digraphs may be more suitable for transaction graphs due to the nature of being attributed and directed. However, there is a lack of exploration of spatial GNNs for digraphs. Several relevant papers did not empirically validate the extension of GNNs to digraphs \cite{gilmer2017neural,journals/corr/LiTBZ15,4700287}, and some do not apply to transaction graphs due to different reasons, such as graph structural constraints \cite{conf/iclr/Thost021}, or aggregation from only out-neighbors \cite{journals/corr/LiTBZ15}. Dir-GNN \cite{conf/log/RossiCGFGB23} is a generic spatial digraph GNN framework, but the use of separate sets of learnable parameters for in- and out-neighbors may be redundant for some domains such as transaction data. Our experimental results show a good model performance despite parameter sharing. However, we design our models within this framework due to its genericness.

% Our empirical results (in Appendix \ref{sec: appendix-parameter-sharing-in-dir-gnn}) show that parameter sharing among GNN aggregation maps for in-neighbors and those for out-neighbors still yields competitive performance in fraud account detection tasks. Therefore, we introduce \textbf{MVGNN} as a specific instance model under a modified Dir-GNN framework. The model shares parameters in the aggregation maps for both types of neighbors. The model can still differentiate messages of each neighbor type by the message combination maps, effectively and efficiently leveraging both the local original view and the local reversed view in a transaction graph.

\begin{figure}[H]
\begin{adjustwidth}{-\extralength}{0cm}
\centering
\includegraphics[width=.98\linewidth]{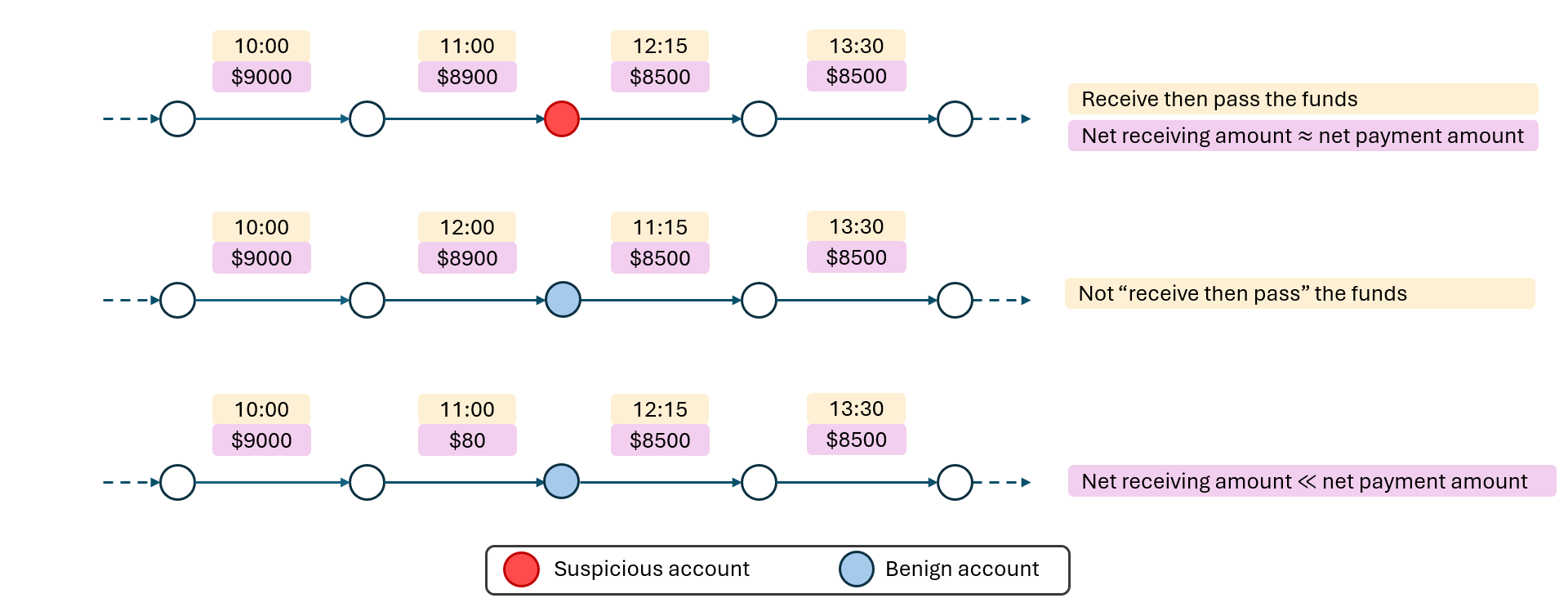}
\end{adjustwidth}
\caption{{{Hypothetical}} %MDPI: Commas are only used for numbers with five or more digits. Please remove them in four-digit numbers, e.g., "1,200" should be "1200".
examples of benign and suspicious accounts in path patterns.}
\label{fig:demo-example-ml-case}
\end{figure}

In addition, specifically for AML detection, identifying suspicious accounts by the relevant transactions depends on cash flow information. Nevertheless, such information may not be effectively captured by SOTA GNNs for digraphs such as DiGCN \cite{conf/nips/TongLSLR020}, MagNet~\cite{conf/nips/ZhangHBPH21}, SigMaNet~\cite{10.1609/aaai.v37i6.25919}, FaberNet \cite{journals/corr/abs-2310-02232}, and Dir-GNN \cite{conf/log/RossiCGFGB23}. As an illustrative example shown in Figure~\ref{fig:demo-example-ml-case}, suppose a series of path-like transactions exist in a transaction graph. The corresponding accounts can be considered suspicious when they serve as temporary repositories for funds~\cite{jfiu2024}.
Identifying these suspicious accounts requires identifying suspicious transactions, and identifying a suspicious transaction requires information from (past) receipt and (future) payment transactions, such as comparing the transaction time and the transaction amounts, to acquire the money flow information. Although stacking GNN layers allows information propagation, the edge-to-edge (transaction-to-transaction) information exchange is indirect and less detailed. Also, the first GNN message passing will be meaningless since raw edge (transaction) attributes are aggregated without comparing with other relevant transactions.
Early propagation of edge (transaction) information and edge updates before the first iteration of GNN message passing can ease the learning of a GNN model for suspicious account detection.

To capture inter-edge interactions, line graphs of edge adjacencies $L(G)$ can be leveraged for edge feature propagation. The line graph $L(G)$ is transformed from the input graph $G$, where $L(G)$ encodes the directed edge adjacency structure of $G$ using the non-backtracking matrix, allowing information to propagate along the directed edges while preserving orientation. We propose {{{LineMVGNN}}%MDPI: Please confirm if the bold should be retained. The highlights below are the same.
} which aggregates and propagates transaction information in the line graph before node (account) feature updates in the input graph $G$.

Our main contributions are as follows:
\begin{itemize}
\item {\textbf{MVGNN}} is introduced as a lightweight yet effective model within the Dir-GNN framework due to its genericness. It supports edge features and considers both in- and out-neighbors in an attributed digraph, such as a transaction graph.
\item {\textbf{{LineMVGNN}}} is proposed, extending MVGNN by utilizing the line graph view of the original graph for the effective propagation of transaction information (edge features in the original graph).
\item Extensive experiments are conducted on the Ethereum phishing transaction network and the financial payment transaction (FPT) dataset.
\end{itemize}

% round 1 : reviewer 3: introduction revision
The remainder of this paper is organized as follows.
Section \ref{sec2} provides an overview of related work in anti-money laundering (AML) and graph neural networks (GNNs) for directed graphs.
Section \ref{sec3} introduces the problem statement and the mathematical framework for our proposed method, including its two-way message passing mechanism and line graph view.
Section \ref{sec4} presents the experimental setup, datasets, and results, comparing LineMVGNN with SOTA methods.
Section \ref{sec5} discusses the limitations of the proposed method and potential future work.
Finally, Section \ref{sec6} concludes the paper with a summary of contributions.

\vspace{-9pt}

\section{Related Work}
\label{sec2}
\subsection{GNNs for Digraphs}
\subsubsection{Spectral Methods}

Recently, researchers have extended spectral convolutions to directed graphs \cite{DBLP:journals/corr/abs-1802-01572,Ma2019SpectralbasedGC,journals/corr/abs-2004-13970}.
In particular, DiGCN \cite{conf/nips/TongLSLR020} uses an approximate digraph Laplacian based on personalized PageRank~\cite{10.14778/1929861.1929864} and incorporates \emph{kth}-order proximity to produce $k$ receptive fields;
MagNet~\cite{conf/nips/ZhangHBPH21} utilizes a complex Hermitian matrix, namely the magnetic Laplacian, to encode both undirected structure and directional information;
SigMaNet \cite{10.1609/aaai.v37i6.25919} unifies the treatment of undirected and directed graphs with arbitrary edge weights by introducing the Sign-Magnetic Laplacian to extend spectral graph convolution theory to graphs with positive and negative weights;
FaberNet \cite{journals/corr/abs-2310-02232} utilizes advanced complex analysis and spectral theory to extend spectral convolutions to directed graphs.   Nevertheless, spectral GNNs might not be appropriate for edge-attributed transaction graphs, because they do not inherently accommodate multi-dimensional edge characteristics and their scalability is restricted by their spectral characteristics.
Typically, full graph propagation is needed during training, with the necessary number of propagations contingent on the degree of the polynomial graph filter.

% However, most of these spectral digraph GNNs share the following limitations:
% (1) They do not naturally support multi-dimensional edge features, which are important for tasks such as money laundering detection in transaction graphs;
% (2) They have reduced interpretability resulting from edge modifications,\footnote{*** unclear why this is needed. see prev footnote in sec1} such as adding/removing self-loops;
% (3) limited scalability owing to their spectral nature.\footnote{*** similar footnote in sec1}

\subsubsection{Spatial Methods}

Though the extension of spatial models to directed graphs is suggested in several classical papers, empirical experiments are not conducted \cite{gilmer2017neural,journals/corr/LiTBZ15,4700287}. GGS-NNs \cite{journals/corr/LiTBZ15} handles directed graphs but only aggregates from out-neighbors.
It neglects in-neighbor information and edge features. Although DAGNN \cite{conf/iclr/Thost021} supports edge features, it is limited to directed acyclic graphs.
Extended from the Message Passing Neural Network (MPNN) framework \cite{gilmer2017neural}, Dir-GNN \cite{conf/log/RossiCGFGB23} is a spatial GNN that accounts for edge directionality by separately aggregating messages from in-neighbors and out-neighbors.  Although the importance of the two types of messages is differentiated by a hyperparameter, Dir-GNN cannot capture the element-wise interaction between these two types of messages. It also lacks efficiency as it doubles the number of parameters to separately aggregate the two types of messages.

Overall, both spectral and spatial methods fail to effectively address the challenges of transaction-based tasks due to their inability to propagate edge-level information effectively.
This limitation is critical for tasks like AML detection, where transaction-level details, such as cash flow patterns, are essential for identifying suspicious activities.
Spectral methods are further hindered by their lack of support for multi-dimensional edge features and scalability issues,
while spatial methods may struggle to capture nuanced interactions between in- and out-neighbors, leading to inefficiencies and suboptimal performance.

\subsection{Edge Feature Learning and Line Graphs}
% \cite{conf/nips/BattagliaPLRK16,Kearnes2016MolecularGC,gilmer2017neural,journals/corr/abs-1710-10903}.
\textls[15]{Combining node and edge feature learning has been suggested by some researchers~\cite{gilmer2017neural,journals/corr/abs-1710-10903}.}
Recently, LGNN \cite{chen2017supervised} operates on line graphs of edge adjacencies, leveraging a non-backtracking operator to enhance performance on community detection tasks. However, the model is restricted to undirected graphs. Multiple works also leverage line graphs for learning edge embeddings, but only for link prediction \cite{Liang2023LineGN,lelgnn2023}.
% \cite{journals/pami/CaiLWJ22,Liang2023LineGN,lelgnn2023}.

Although node and edge feature information usually complement each other, edge feature propagation and learning for node classification tasks remain relatively unexplored. To incorporate multi-dimensional edge features for node feature updates, Zhang et al. \cite{10004977} proposes a graph representation learning framework that generates node embeddings by aggregating {{local}} edge embeddings. Nevertheless, the learned edge embeddings do not effectively capture the interactions between edges, since the edge embeddings are learned from the concatenation of the edge features and the features of the corresponding end~nodes.

To capture inter-edge interactions and enhance node embeddings, line graphs of edge adjacencies can be leveraged, but this remains relatively unexplored for node classification tasks. CensNet \cite{conf/ijcai/JiangJL19} co-embeds nodes and edges by switching their roles using line graphs, but is limited to undirected graphs without parallel edges because of the use of spectral graph convolution. LineGCL \cite{10.1016/j.jksuci.2024.102011} transforms the original graph into a line graph to enhance the representation of edge information and facilitate the learning of node features by contrastive learning, but it assumes no edge features in the original graph.

% IN \cite{conf/nips/BattagliaPLRK16} adopts edge updates to reason about physical systems using attributed multigraphs.
% LaundroGraph \cite{10.1145/3533271.3561727} represents edges as nodes and implements a GNN on the line graph.

\section{Proposed Method}
\label{sec3}
\subsection{Problem Statement}
Consider a directed transaction graph $G=(\mathcal{V},\mathcal{E})$, where $\mathcal{V}$ is the set of nodes $\{v_1, \dots, v_n\}$ and $\mathcal{E}$ is the set of edges $\{e_1, \dots, e_m\}$. Each node represents an account and each edge represents a transaction. An edge's source node and destination node represent the payer and payee of the corresponding transaction, respectively. Each node $v$ is attributed with $x_v$ and each edge is attributed with $e_{vw}$. Each node is either licit or illicit, and is represented by 0 and 1, respectively. A subset of nodes are unlabeled. We aim to learn a model that predicts the label $y_v$ of each unlabeled node $v$.

\subsection{Two-Way Message Passing}

\begin{figure}[H]
\hspace{-6pt}  \includegraphics[width=1\linewidth]{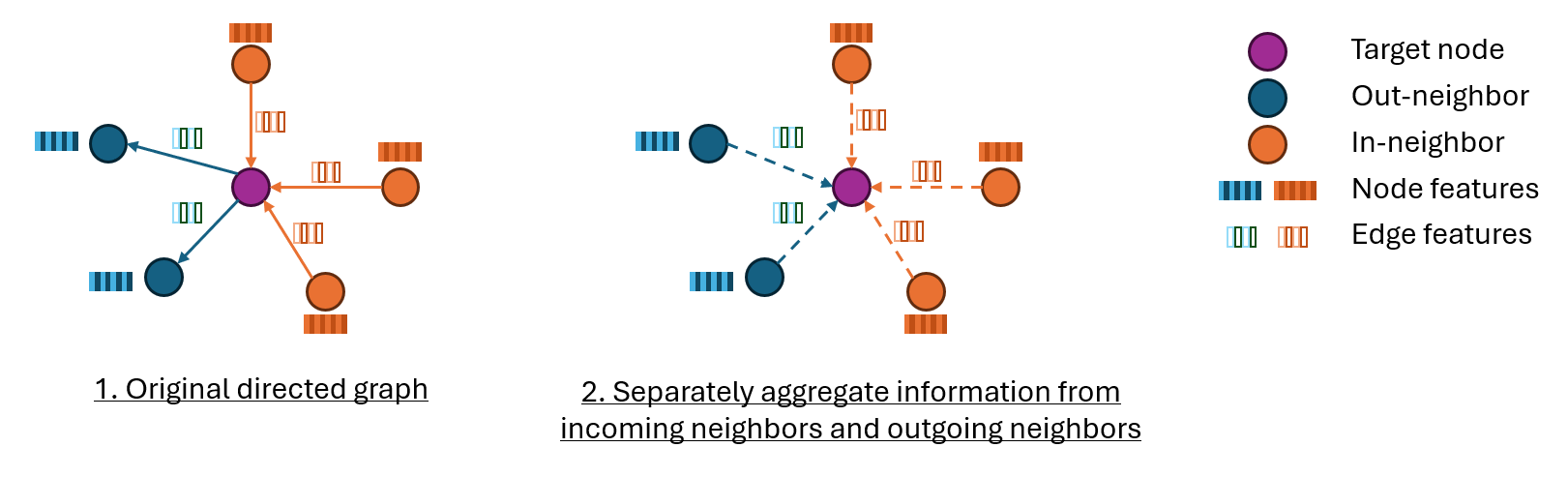}
\caption{{{Visual}} %MDPI: Please confirm if different color shape need explanation?
illustration of two-way message passing in MVGNN.}
\label{fig:mvgnn}
\end{figure}

Information from both the in- and out-neighbors and edges is important for node
classification. In a digraph, however, messages from the out-neighbors and edges are usually ignored in traditional GNNs such as GCN \cite{Kipf:2016tc}, GraphSAGE \cite{HamiltonYL17}, GIN \cite{journals/corr/abs-1810-00826}, etc. In addition, edge
attributes are present in some real-world data, and they can be significantly more
informative than node attributes. Therefore, we build our simple {\textbf{{MVGNN}} (\underline{{M}}ulti-\underline{{V}}iew \underline{{G}}raph \underline{{N}}eural \underline{{N}}etwork)} model within the Dir-GNN framework
\cite{conf/log/RossiCGFGB23}.
{As illustrated in Figure \ref{fig:mvgnn},} the model considers messages from both
in-neighbors and out-neighbors.
Our MVGNN differs from Dir-GNN by (1) parameter sharing between message aggregation functions for in- and out-neighbors, and (2) the message combination function for the two types of messages.
{Details are provided below.}

\vspace{-6pt}

At the $(l+1)$-th MVGNN layer,
% similar to equation \ref{eq: dir-gnn msg in} and \ref{eq: dir-gnn msg out},
aggregation of messages $m_{v_{in}}^{(l+1)}$ and
$m_{v_{out}}^{(l+1)}$ can be expressed by the following:
\begin{eqnarray}
m_{v_{in}}^{(l+1)} = \sum_{w_k \in N_{in}(v)}M_{in}^{(l+1)}\left( [ h_{w}^{(l)} \Vert e_{vw} ] \right),
\label{eq: msg in} \\
m_{v_{out}}^{(l+1)} = \sum_{w_k \in N_{out}(v)}M_{out}^{(l+1)}\left( [ h_{w}^{(l)} \Vert e_{vw} ] \right),
\label{eq: msg out}
\end{eqnarray}
where $M_{in}^{(l)}$ and $M_{out}^{(l)}$ are fully connected layers
with ReLU activation, $N_{in}(v)$ and $N_{out}(v)$ denote the in- and out-neighbors of node $v$, $e_{vw}$ denotes the features of an edge between node $v$ and $w$, and $\Vert$ denotes concatenation.

% similar to equation \ref{eq: dir-gnn com-general},
The node embeddings are updated with these messages by the equation below:
\begin{equation}
h_v^{(l+1)} = U^{(l+1)}\left(h_v^{(l)}, C^{(l+1)}\left(m_{v_{in}}^{(l+1)}, m_{v_{out}}^{(l+1)}\right) \right),
\end{equation}
where the vertex update function $U^{(l+1)}$ is a fully connected layer with ReLU activation. For the message combination function $C^{(l+1)}$, we experiment with 2 different methods. For brevity and clarity, superscript is omitted, i.e., $C^{(l+1)}$, $m_{v_{in}}^{(l+1)}$, and $m_{v_{out}}^{(l+1)}$ will become $C$, $m_{v_{in}}$, and $m_{v_{out}}$ respectively.

{\textbf{{Combine by Weighted Sum.}}} To differentiate the importance of messages from in-neighbors and out-neighbors, similarly to \cite{conf/log/RossiCGFGB23}, we propose a variant {\textbf{{MVGNN-}\textsl{{add}}}}, which multiplies weight scalars with $m_{v_{in}}$ and $m_{v_{out}}$:
\begin{equation}
\label{eq: h combine by add}
C \left(m_{v_{in}}^{(l+1)}, m_{v_{out}}^{(l+1)}\right) = \alpha_{l+1} \, m_{v_{in}}^{(l+1)} + (1-\alpha_{l+1}) \, m_{v_{out}}^{(l+1)},
\end{equation}
where $\alpha_{l+1} \in \mathbb{R}$ is a {{learnable}} scalar (instead of a hyperparameter as in \cite{conf/log/RossiCGFGB23}).

{\textbf{{Combine by Concatenation.}}} To better model the element-wise interaction between messages from these two types of neighbors, {\textbf{{MVGNN-}\textsl{{cat}
}}} is proposed (instead of using weighted sum as in \cite{conf/log/RossiCGFGB23}). The two types of messages are concatenated and then passed to a linear layer $F^{(l+1)}$:

% round 1 : reviewer 2 : typo (for the use of concatenation symbol)
\begin{equation}
\label{eq: h combine by cat}
C \left(m_{v_{in}}^{(l+1)}, m_{v_{out}}^{(l+1)}\right) =
F^{(l+1)}\left(
[ m_{v_{in}}^{(l+1)} \ \Vert \ m_{v_{out}}^{(l+1)} ]
\right).
\end{equation}

To boost model efficiency, we share parameters for aggregation maps, message combination maps, and vertex update functions for both types of neighbors in our experiments. In other words, the primary power of distinguishing between in- and out-messages lies in the learnable parameter $\alpha$, or the linear layer $F$. In experiments, this proposed model remains competitive.

In short, at a high level, the $(l+1)$-th MVGNN layer can be expressed as follows:
\begin{equation}
\label{eq:mvgnn-general-eq}
h_v^{(l+1)} = \text{MVGNNLayer}^{(l+1)}\left(
\{h_{u}^{(l)}, h_{w}^{(l)}, e_{vu}^{(l)}, e_{vw}^{(l)} \ \vert \
u \in \mathcal{N}_{in}(v),
\ w \in \mathcal{N}_{out}(v)\}
\right),
\end{equation}
% \begin{equation}
%     \label{eq:mvgnn-general-eq}
%     h_v^{(l+1)} = \text{MVGNNLayer}^{(l+1)}(\{h_{u}^{(l)}, h_{w}^{(l)}, e_{vu}^{(l)}, e_{vw}^{(l)}\}),
% \end{equation}
% for all $u \in \mathcal{N}_{in}(v), w \in \mathcal{N}_{out}(v)$,
where MVGNNLayer $= \{\text{MVGNNLayer-add}$, and $\text{MVGNNLayer-cat}\}$, $e_{vu}^{(l)}$ and $e_{vw}^{(l)}$ are edge embeddings. Since there are no edge updates, $e_{vu}^{(l)} = e_{vu}$ and $e_{vw}^{(l)} = e_{vw}$. ``MVGNNLayer-add'' denotes message combination by weighted sum defined in Equation (\ref{eq: h combine by add}), while ``MVGNNLayer-cat'' denotes message combination by concatenation followed by a linear~layer.

% \begin{equation}
%     \label{eq:mvgnn-general-eq}
%     \begin{aligned}
%         h_v^{(l+1)} = \text{MVGNNLayer}^{(l+1)}(\{h_{u}^{(l)}, h_{w}^{(l)}, e_{vu}^{(l)}, e_{vw}^{(l)}\}: \\
%         \forall u \in \mathcal{N}_{in}(v), w \in \mathcal{N}_{out}(v)), \\
%         \text{MVGNNLayer} = \{\text{MVGNNLayer-add}, \text{MVGNNLayer-cat}\}
%     \end{aligned}
% \end{equation}

To prevent over-smoothing neighbors’ information, similar to \cite{conf/ijcai/GongWSLN0P23}, personalized PageRank-based aggregation mechanism of GNN \cite{conf/iclr/ChienP0M21} is adopted to obtain the final aggregated embedding $z_v$ of node $v$:
\begin{equation}
\label{eq:agg-node-emb-for-pred}
\begin{aligned}
z_v = \sum_{l=1}^{L-1} \beta_{l} h_{v}^{(l)} + \left(1-\sum_{l=1}^{L-1} \beta_{l} \right) \times h_{v}^{(L)},
\end{aligned}
\end{equation}
where $\beta_{l} \in \mathbb{R}$ is a \textit{learnable} parameter. Subsequently, a fully connected layer maps each $z_v$ to a prediction vector $y_v$.

% To aggregate messages regardless of edge direction, one naive way is to add a real-valued attribute to the nodes and edges to indicate whether they are in- or out-neighbours (for example, 1 for in-neighbours, -1 for out-neighbours) [ === WIP === : cite some relevant papers ]. Then, perform message passing from all neighbors. However, this approach is not expressive enough because the interactions between each attribute may not be similar. Sharing the same graph convolutional layer assumes such similarity. Even if the interaction is similar, this approach can hardly differentiate the importance (weights) of the aggregated messages from the in-neighbors against those from the out-neighbors. [ === WIP === : proof]

\subsection{Line Graph View}

Whether a transaction is illicit also depends on the previous transactions due to the significance of the money flow. To facilitate capturing this information, we leverage line graphs to perform edge feature propagation. To the best of our knowledge, few studies have leveraged line graphs in anti-money laundering. One example is LaundroGraph \cite{10.1145/3533271.3561727} which represents edges as nodes and implements a GNN on the line graph.

The original transaction graph $G$ is transformed into a line graph $G' = \mathcal{L}(G)$, in which each node $t(v,w) \in G'$ corresponds to edge $e_{vw} \in G$ (transaction from $w$ to $v$). In the line graph $G'$, a directed edge $e'_{t(v,w) t(u,s)}$ exists if $u=w$ (i.e., the {pay{ee}} $u$ of transaction $t(u,s)$ is the {pay{er}} $w$ of transaction $t(v,w)$).

\begin{figure}[H]
\centering
\begin{adjustwidth}{-\extralength}{0cm}
\includegraphics[width=1\linewidth]{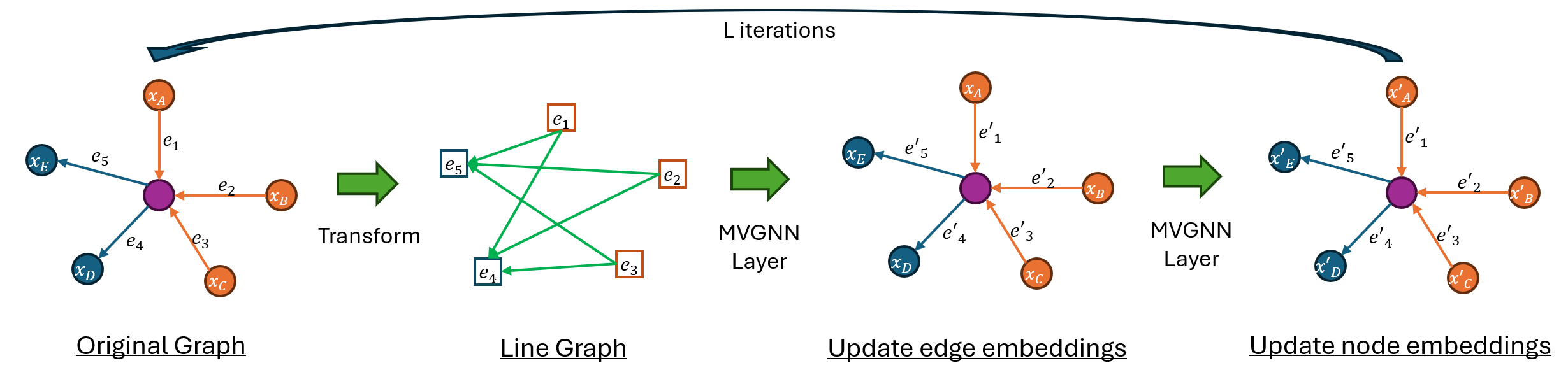}
\end{adjustwidth}
\caption{Visual illustration of LineMVGNN.}
\label{fig:linemvgnn}
\end{figure}

Figure \ref{fig:linemvgnn} and Algorithm \ref{alg: LineMVGNN} show how our proposed {\textbf{{LineMVGNN}}} leverages the line graph view and propagates edge features. In the line graph $G'$, edge feature propagation is performed with $G'$ before each message passing in $G$.
LineMVGNN uses separate MVGNN layers for the original graph $G$ and the line graph $G'$, respectively, as defined in Equation~(\ref{eq:mvgnn-general-eq}).
To seamlessly involve the line graph view in every message passing in the original graph $G$, inspired by the cross-stitch networks \cite{Misra_2016_CVPR}, we update the edge features with $G'$ before each message propagation in $G$ as shown from line \ref{algstep:linemvgnn-step-1} to \ref{algstep:linemvgnn-step-3} in Algorithm~\ref{alg: LineMVGNN}. Residual connection~\cite{He_2016_CVPR} is adopted for edge embedding updates, as shown in line \ref{algstep:linemvgnn-step-2}. Note that a dummy edge feature $[\mathbf{{{1}}}]$ can be used for models requiring edge features, which are usually absent in $G'$. With the use of line graphs, edge features can be effectively propagated. In other words, information about a particular transaction can be propagated to the next transactions, capturing money flow.

\begin{algorithm}[H]
\caption{LineMVGNN}
\label{alg: LineMVGNN}
\textbf{Input}:
Graph $G = (\mathcal{V},
\mathcal{E})$;
input node features $\{h_{v}^{(0)}, \ \forall \ v \in \mathcal{V}\}$;
input edge features $\{e_{vw}^{(0)}, \ \forall \ v \in \mathcal{V}\ , w \in \mathcal{N}_{out}(v)\}$;
model depth $L$ \\
\textbf{Parameter}: $\{\text{MVGNNLayer}_{G}^{(l)}, \text{MVGNNLayer}_{G'}^{(l)}\} : \forall l \in L$\\
\textbf{Output}: Vector $z_v, \forall v \in \mathcal{V}$
\begin{algorithmic}[1] %[1] enables line numbers
\STATE $G' \leftarrow \mathcal{L}(G)$
\STATE Initialize in $G$: $h_{v}^{(0)} \leftarrow x_{v}$.
\STATE Initialize in $G'$: $t_{v,w}^{(0)} \leftarrow e_{vw}^{(0)}$.
\FOR{$l = 0$ \TO $L$}
\STATE \label{algstep:linemvgnn-step-1} With $t^{(l)}(v,w)$ in $G'$, apply $\text{MVGNNLayer}_{G'}^{(l+1)}$ and obtain $t^{(l+1)}(v,w)$ by \mbox{Equation~(\ref{eq:mvgnn-general-eq}).}
\STATE \label{algstep:linemvgnn-step-2} $e_{vw}^{(l)} \leftarrow t^{(l+1)}(v,w)$
\STATE \label{algstep:linemvgnn-step-3} With $e_{vw}^{(l)}$ in $G$, apply $\text{MVGNNLayer}_{G}^{(l+1)}$ and obtain $h_{v}^{(l+1)}$ by Equation (\ref{eq:mvgnn-general-eq}).
\ENDFOR
\STATE $z_v \leftarrow$ Equation (\ref{eq:agg-node-emb-for-pred})
\STATE \textbf{return} $z_v$
\end{algorithmic}
\end{algorithm}

The LineMVGNN model is further divided into 2 variants: {\textbf{{LineMVGNN-}\textsl{{add}}}} if using weighted sum for combining messages as defined in Equation (\ref{eq: h combine by add}), or {\textbf{{LineMVGNN-}\textsl{{cat}}}} which concatenates message vectors followed by a linear layer as defined in Equation (\ref{eq: h combine by cat}).

% % SKIPPED :
% Similar to equation \ref{eq:mvgnn-general-eq}, the edge feature $e_{vw}^{(l+1)} \in G$ is updated by:
% \begin{equation}
%     \begin{aligned}
%             e_{vw}^{(l+1)} = e_{vw}^{(l)} + TwoWayMsgPass( \\
%             h_{v'_{in}}^{(l)}, h_{v'_{out}}^{(l)}, \mathcal{N}_{in}(v'), \mathcal{N}_{out}(v')).
%     \end{aligned}
% \end{equation}

% % SKIPPED :
% Similar to equation \ref{eq:mvgnn-general-eq}, the edge feature $e_{vw}^{(l+1)} \in G$ is updated by:
% \begin{equation}
%     \begin{aligned}
%             e_{vw}^{(l+1)} = e_{vw}^{(l)} + TwoWayMsgPass( \\
%             h_{v'_{in}}^{(l)}, h_{v'_{out}}^{(l)}, \mathcal{N}_{in}(v'), \mathcal{N}_{out}(v')).
%     \end{aligned}
% \end{equation}

% instead of setting $e_{vw}^{(l+1)} = e_{vw}^{(0)}$.

% round 1 : reviewer 1 : addressing computational complexity

% < worst-case complexity : complete graph >
% Line graph construction : O(|E|^2)
% Edge information propagation on L(G) : O(|E|^2)
% Node propagation on G : O(|E|)
% Overall : O(|E|^2)
\subsection{{Computational Complexity}}
{
The overall complexity of this propagation scheme is dominated by the edge propagation on the line graph $G'$, which has a complexity of $\mathcal{O}(L|\mathcal{E}|^2)$ for $L$ model layers,
and the line graph construction,
which has a complexity of $\mathcal{O}(|\mathcal{E}|^2)$ in the worst case, such as in a complete graph.
The propagation scheme can be optimized by avoiding explicit line graph construction.
Edge propagation can be performed directly on the original graph $G$, as shown in Algorithm \ref{alg: optimized-LineMVGNN}.
In this refined LineMVGNN, edges in $G$ are treated as first-class entities.
For each edge $e_{vw} \forall (v, w) \in \mathcal{E}$, messages are aggregated from neighboring edges that share a common node.
With a sparse graph, this reduces the asymptotic computational complexity to $\mathcal{O}(L|\mathcal{E}|)$, which is the same with GCN \cite{Kipf:2016tc}.
}

\begin{algorithm}[H]
\caption{Refined LineMVGNN (Without Explicit Line Graph Construction)}
\label{alg: optimized-LineMVGNN}
\textbf{Input}:
Graph $G = (\mathcal{V},
\mathcal{E})$;
input node features $\{h_{v}^{(0)}, \ \forall \ v \in \mathcal{V}\}$;
input edge features $\{e_{vw}, \ \forall \ v \in \mathcal{V}\ , w \in \mathcal{N}_{out}(v)\}$;
model depth $L$ \\
\textbf{Parameter}: $\{\text{MVGNNLayer}_{\text{node}}^{(l)} \ , \ \text{MVGNNLayer}_{\text{edge}}^{(l)}\} : \forall \ l \in L$\\
\textbf{Output}: Vector $z_v, \ \forall \ v \in \mathcal{V}$
\begin{algorithmic}[1] %[1] enables line numbers
\STATE $h_{v}^{(0)} \leftarrow x_{v}$.
% \STATE $e_{vw}^{(-1)} \leftarrow e_{vw}^{(0)}$.
\FOR{$l = 0$ \TO $L$}
\STATE \# Edge Feature Propagation
\STATE Scan for $ N_{in}(e_{vw})$ and $ N_{out}(e_{vw})$.
\STATE $e_{vw}^{l} \leftarrow$ apply $\text{MVGNNLayer}_{\text{edge}}^{(l)}$ by Equation (\ref{eq:mvgnn-general-eq}), treating $e_{vw}$ as the first-class entities.

\STATE \# Node Feature Propagation
\STATE Scan for $N_{in}(v)$ and $N_{out}(v)$.
\STATE $h_{v}^{(l+1)} \leftarrow$ apply $\text{MVGNNLayer}_{\text{node}}^{(l+1)}$ by Equation (\ref{eq:mvgnn-general-eq}).
\ENDFOR
\STATE $z_v \leftarrow$ Equation (\ref{eq:agg-node-emb-for-pred})
\STATE \textbf{return} $z_v$
\end{algorithmic}
\end{algorithm}

\section{Experiments}
\label{sec4}
% In this section, we first describe the datasets for our experiments, and then compare our proposed models with other state-of-the-art baselines. Finally, ablation studies are performed on the variants of our proposed models.

\subsection{Datasets}
Two datasets are used to evaluate LineMVGNN in the classification of benign nodes against illicit counterparts.

\subsubsection{Ethereum (ETH) Datasets} Ethereum Phishing Transaction Network
data is a real public transaction graph dataset available on Kaggle.
Each node represents an address, and each edge represents a transaction. The nodes
are labeled as either phishing or not. Each edge contains two attributes: the
balance and the timestamp of the transaction.
To address the node class imbalance and huge graph size, we adopt graph sampling strategies
similar to \cite{kanezashi2022ethereum,9184813}, creating two data subsets with different subgraph sizes, namely {{ETH-Small}} (12,484 nodes and \mbox{762,443 edges}) and {{ETH-Large}} (30,757 nodes and 1,298,315 edges).
Since no node attributes are provided in the original dataset, we further derive variant data subsets by adding structural node features (SNFs): in-degrees and out-degrees, yielding a total of four data subsets: {{ETH-Small (w/ SNF)}}, {{ETH-Large (w/ SNF)}}, {{ETH-Small (w/o SNF)}}, and {{ETH-Large (w/o SNF)}}.
We adopt a semi-supervised transductive learning setting to classify nodes as illicit (fraud) nodes or benign, and treat all non-central nodes as unlabeled.
Nodes of each extracted subgraph are randomly split into training, validation, and test sets at a ratio of 60\%:20\%:20\%.

% First, all 1,165 fraud nodes are selected, and 1,165 normal nodes are randomly chosen. These nodes
% are referred to as "central nodes" in this paper. Then, all first-hop in- and
% out-neighbors, which will be referred to as "root nodes", are extracted. To
% extract a subgraph, we adopt two strategies: (1) random walks of length 3 are applied and initiated from the root nodes to extract a small subgraph, namely \textsl{ETH-Small}; (2) extract all k-hop in- and out-neighbors from the root nodes, forming a large subgraph, namely \textsl{ETH-Large}. \textit{ETH-Small} contains 12,484 nodes and 762,443 edges. \textit{ETH-Large} contains 30,757 nodes and 1,298,315 edges. For both extracted subgraphs, min-max scaling is applied to transaction balance and timestamps.

% Since no node attributes are provided in the original dataset, we added structural node features (SNFs): in-degrees and out-degrees. An integer "1" is also added as an additional node attribute. These datasets are referred to \textbf{ETH-Small (w/ SNF)} and \textbf{ETH-Large (w/ SNF)}. To comprehensively reflect model performances, two variant datasets \textbf{ETH-Small (w/o SNF)} and \textbf{ETH-Large (w/o SNF)} are derived by masking all SNF. We adopt a semi-supervised transductive learning setting to classify nodes as illicit (fraud) nodes or benign and treat all non-central nodes as unlabeled. Nodes of each extracted subgraph are randomly split into training, validation, and test sets at a ratio of 60\%:20\%:20\%.

\subsubsection{Financial Payment Transaction (FPT) Dataset}

Provided by our industry partner, this transaction dataset contains e-wallet payment transaction data from January 2022.  After data preprocessing, a transaction graph
is constructed for each day, where accounts and transactions are represented by
nodes and edges, respectively.
No SNFs are added. As we assume that the real data contain no anomalous money laundering patterns, synthetic anomalies are injected into the graphs, following the injection strategy from \cite{Elliott2019AnomalyDI}.
The proportion of illicit (money laundering) nodes constitutes roughly one-third of the total node count.
On average, there are 1,048,512 nodes and 1,092,895 edges in the graphs for each day.
A supervised transductive learning setting is used to classify nodes as either illicit (money laundering) or benign.
The transaction graphs for the whole month are chronologically split into training, validation, and test sets at a ratio of 60\%:20\%:20\%.
{For more details, please refer to Appendix \ref{sec: FPT Dataset}.}

\subsection{Compared Methods and Evaluation Metrics}
The following baseline GNNs are compared with our proposed methods:
(1) {{Non-digraph GNNs}} include {{GCN}} \cite{Kipf:2016tc}, {{GraphSAGE}}, \cite{HamiltonYL17}, {{MPNN}} \cite{gilmer2017neural}, {{GIN}} \cite{journals/corr/abs-1810-00826}, {{PNA}} \cite{CorsoCBLV20}, and {{EGAT}}~\cite{10.1093/bib/bbab371};
(2) {{Digraph GNNs}} includes {{DiGCN}} \cite{conf/nips/TongLSLR020}, {{MagNet}} \cite{conf/nips/ZhangHBPH21}, {{SigMaNet}} \cite{10.1609/aaai.v37i6.25919}, {{FaberNet}}~\cite{journals/corr/abs-2310-02232}, and {{Dir-GCN}} and {{Dir-GAT}
} \cite{conf/log/RossiCGFGB23}.

Since jumping knowledge \cite{Xu:2018vn} by concatenation (``cat'') and by max-pooling (``max'') are applied in FaberNet, Dir-GCN, and Dir-GAT, we build the corresponding variant models, namely {{FaberNet (cat)}}, {{FaberNet (cat)}}, {{Dir-GCN (cat)}%MDPI: Please confirm if special font is necessary？please check all.
}, {{Dir-GCN (max)}}, {{Dir-GAT (max)}}, and {{Dir-GAT (cat)}}. For models that do not naturally support edge features, we concatenate them with node features.
% For more model details and implementation details, please refer to Appendix \ref{sec: appendix-compared-methods} and \ref{sec: appendix-implementation-details} respectively.

Since some datasets are very imbalanced, the F1 score for the illicit class is used for model performance evaluation, which is similar to what is used in real-world scenarios.

\subsection{Results}

Table \ref{table: main results - f1} summarizes the F1 scores of the illicit class across five datasets for all models. Our LineMVGNN model, especially the variant model LineMVGNN-\textsl{cat}, achieves state-of-the-art results on nearly all datasets.

\begin{table}[H]\setlength{\tabcolsep}{2.72mm}
\footnotesize % \small
% Summary_all_results_impl.xlsx
\caption{F1 scores of the illicit class. For each dataset, the highest F1 score is \textbf{\underline{bold and underlined}}, and the 1st runner-up is \underline{underlined}. ``OOM'' indicates out of memory.}
\label{table: main results - f1}
% \begin{adjustwidth}{-\extralength}{0cm}
\begin{tabular}{llccccc}
\toprule
\multirow{2}{*}{\textbf{Category}\vspace{-5pt}} & \multirow{2}{*}{\textbf{Methods}\vspace{-5pt}} & \multicolumn{2}{c}{\textbf{ETH-Small}} & \multicolumn{2}{c}{\textbf{ETH-Large}} & \textbf{FPT} \\
\cmidrule{3-7}
& & \textbf{w/ SNF} & \textbf{w/o SNF} & \textbf{w/ SNF} & \textbf{w/o SNF} & \textbf{w/o SNF} \\

\midrule
\multirow{6}{*}{Non-Digraph GNNs} & GCN & 0.8770 & 0.8998 & 0.9068 & 0.9072 & 0.8817\\    % gcn_nowmsg_normboth ; 0.8770 , 0.9068, 0.9072 (2nd)
& GraphSAGE & 0.8752 & 0.6705 & 0.8984 & 0.6705 & 0.8802\\    % 0.8752 (2nd)
& MPNN & 0.7857 & 0.8912 & 0.8854 & 0.9087 & OOM \\
& GIN & 0.9055 & 0.8954 & 0.9117 & 0.8950 & 0.8802 \\
& PNA & 0.9352 & 0.9105 & 0.9130 & 0.9249 & OOM \\
& EGAT & 0.8916 & 0.6705 & 0.9195 & 0.6705 & OOM \\
\midrule

%  edgedigcn_ib	0.8192	0.8055	0.865	0.829
\multirow{9}{*}{Digraph GNNs} & DiGCN & 0.8192 &  0.8055 & 0.8650 & 0.8290 & OOM \\    % DiGCN on FPT: OOM when get_appr_directed_adj(), even on CPU with over 1500 GB memory

% & MagNet & 0.9342 & 0.9087 & \underline{0.9561} & \underline{\textbf{0.9571}} & 0.9616 \\    % edgemagnet_efeat2eweight for ETH ;  edgemagnet_catsrcfeat for FPT

& MagNet & 0.9009 & 0.9012 & 0.9330 & 0.9354 & 0.9616 \\  % edgemagnet_preprocess for ETH ;  edgemagnet_catsrcfeat for FPT

& SigMaNet & 0.8072 & 0.8319 & 0.8018 & 0.8300 & 0.5033 \\

& FaberNet (cat) & 0.9352	& 0.9393 & \underline{0.9476} & 0.9451 & 0.9934 \\    % 0.9352 : best lr but seed 0 (2nd) ; 0.9451 : best lr but seed 42 (4th)
& FaberNet (max) & 0.9336 & 0.9376 & 0.9381 & \underline{0.9460} & \underline{0.9945} \\    % 0.9376 : best lr but seed 0 (2nd)

% Dir-GCN (cat) & 0.9240 & 0.9023 & 0.9039 & 0.9120 & 0.6577 \\    % To be replaced : because norm only the nfeats
% Dir-GCN (max) & 0.8826 & 0.8427 & 0.9121 & 0.8230 & 0.6576 \\     % To be replaced : because norm only the nfeats

& Dir-GCN (cat) & 0.9240 & 0.8987 & 0.9168 & 0.9188 & 0.6402 \\    % version: normalize the concatenated node and edge features
& Dir-GCN (max) & 0.8577 & 0.9000 & 0.8598 & {0.9207} & 0.6402 \\    % version: normalize the concatenated node and edge features

& Dir-GAT (cat) & 0.8831 & 0.6705 & 0.8769 & 0.6705 & 0.9768 \\
& Dir-GAT (max) & 0.7958 & 0.6705 & 0.8515 & 0.6705 & 0.9908 \\

% Dir-GCN (cat, shared) & 0.9091 & 0.9264 & 0.9243 & 0.9197 & - \\
% Dir-GCN (max, shared) & 0.9119 & 0.8733 & 0.9192 & 0.9039 & - \\
% Dir-GAT (cat, shared) & 0.8827 & 0.6705 & 0.8719 & 0.6705 & - \\
% Dir-GAT (max, shared) & 0.8396 & 0.6705 & 0.8632 & 0.6705 & - \\
\midrule
\multirow{4}{*}{Our Digraph GNNs} & MVGNN-{{add}} & 0.9231 & 0.9333 & 0.9300 & 0.9365 & 0.9821\\
& MVGNN-{{cat}} & 0.9331 & 0.9301 & 0.9439 & 0.9394 & 0.9858 \\
& LineMVGNN-{{add}} & \underline{0.9362} & \underline{0.9407} & \underline{\textbf{0.9598}} & 0.9048 & 0.9905 \\
& LineMVGNN-{{cat}} & \underline{\textbf{0.9441}} & \underline{\textbf{0.9455}} & 0.9394 & \underline{\textbf{0.9565}} & \underline{\textbf{0.9954}} \\
\bottomrule
\end{tabular}
% \end{adjustwidth}
\end{table}

Compared to non-digraph GNN baselines, LineMVGNN beats the competing methods by an average of 9.68\% across all datasets. For each dataset, the improvement is at least 0.89\%, 3.50\%, 4.03\%, 3.16\%, and	11.37\% on the ETH-Small (w/ SNF), ETH-Small (w/o SNF), ETH-Large (w/ SNF), ETH-Large (w/o SNF), and FPT datasets, respectively. Compared to digraph GNN baselines, LineMVGNN improves the prediction illicit F1 scores by an average of 10.79\% across all datasets. For each dataset, the increment is at least 0.89\%, 0.62\%, 1.22\%, 1.05\%, and 0.09\%, respectively. It is noteworthy that our LineMVGNN model achieves over 99\% in the FPT dataset without SNFs. This shows the effectiveness of leveraging both the line graph view and the reverse view for node classification tasks (illicit account detection). In addition, even with shared parameters among GNN aggregation maps for in-neighbors and those for out-neighbors, the performance of our MVGNN variant models matches with other competing digraph GNN methods. This confirms that high model efficiency and expressiveness were achieved.

\subsection{{Discussion}}

\subsubsection{Effect of Different Views}
As shown in Table \ref{table: ablation - effect of the views and structural node features}, regardless of the presence of structural node features (SNFs), the performance in the illicit node detection drops when we remove the line graph view for LineMVGNN-{{cat}} and LineMVGNN-{{add}}. A more significant drop is caused when we remove the reversed view {(}%MDPI: Please add the corresponding ).
i.e., only aggregating messages from in-neighbors{{)}}.
This reflects the contributions of each view in the illicit node classification tasks.

\begin{table}[H]\setlength{\tabcolsep}{2.455mm}
\small
% Summary_all_results_impl.xlsx
% Results_GNN_FPS_LSTM.xlsx (in LSCM SharePoint)
\caption{Illicit F1 in the ablation experiments. ``TWMP'' and ``LGV'' stand for ``two-way message passing'' and ``line graph view'', respectively.}
\label{table: ablation - effect of the views and structural node features}
\begin{tabular}{llcccccc}
\toprule
\multirow{2}{*}{\textbf{Method}\vspace{-5pt}} & \multirow{2}{*}{\textbf{Components}\vspace{-5pt}} & \multicolumn{2}{c}{\textbf{ETH-Small}} & \multicolumn{2}{c}{\textbf{ETH-Large}} & \textbf{FPT} \\
\cmidrule{3-7}
& & \textbf{w/ SNF} & \textbf{w/o SNF} & \textbf{w/ SNF} & \textbf{w/o SNF} & \textbf{w/o SNF} \\

\midrule
LineMVGNN-{{cat}} & TWMP + LGV & 0.9441 & 0.9455 & 0.9394 & 0.9565 & 0.9954 \\    % copy LineMVGNN-\textsl{cat} results from main results
& {\quad {- LGV}}
%MDPI: PLease confirm if  indent is necessary?
%author: yes, because it is progressively removing the components one-by-one.
& 0.9331 & 0.9301 & 0.9439 & 0.9394 & 0.9858 \\   % copy MVGNN-\textsl{cat} results from main results
& {\qquad {- TWMP}} & 0.9009 & 0.8922	& 0.9042 & 0.9031 & 0.8188 \\   % copy "gcn" (neither "gcn_nowmsg_normboth" nor "gcn_nowmsg") results from excel file
\midrule

LineMVGNN-{{add}} & TWMP + LGV & 0.9362 & 0.9407 & 0.9598 & 0.9048 & 0.9905 \\    % copy LineMVGNN-\textsl{add} results from main results
& {\quad {- LGV}} & 0.9231 & 0.9333 & 0.9300 & 0.9365 & 0.9821\\   % copy MVGNN-\textsl{add} results from main results
& {\qquad {- TWMP}} & 0.9009 & 0.8922	& 0.9042 & 0.9031 & 0.8188 \\   % copy "gcn" (neither "gcn_nowmsg_normboth" nor "gcn_nowmsg") results from excel file
\bottomrule
\end{tabular}
\end{table}

\subsubsection{Effect of SNF}
As shown in Table \ref{table: ablation - effect of the views and structural node features}, regardless of the presence of different views, our LineMVGNN-{{cat}} model performs robustly when SNF (including node in-degrees and out-degrees) is masked in the ETH-Small and ETH-Large datasets. In addition, when LineMVGNN-{{cat}} is compared with LineMVGNN-{{add}}, message combination by concatenation is superior to the weighted sum method. Without SNF, the performance of LineMVGNN-{{add}} drops a little in various scenarios over Eth-Small and Eth-Large. Also, the performance of our LineMVGNN-{{cat}} model is less sensitive to the absence of SNF than LineMVGNN-{{add}}. This reflects the superiority of message combination by concatenation following a linear transformation over message combination by weighted sum.

\subsubsection{{Effect of Parameter Sharing}}

Our LineMVGNN variants extend from our MVGNN model, which is designed within the Dir-GNN framework due to its versatility.
A key distinction between the MVGNN and Dir-GNN models lies in their parameter sharing approach.
For Dir-GNN models, two independent sets of learnable parameters for $\text{M}_{in}^{(l)}$ and $\text{M}_{out}^{(l)}$ are used, theoretically enhancing expressiveness but doubling the number of parameters for each Dir-GNN layer.
In contrast, our MVGNN models employ parameter sharing, creating a lighter yet effective~model.

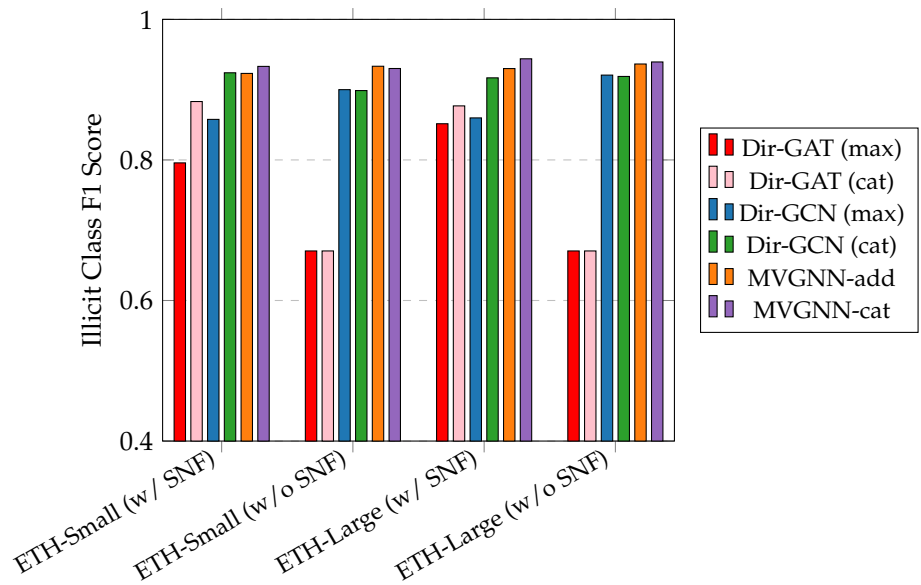
\begin{figure}[H]
\hspace{-10pt}  \begin{tikzpicture}
\begin{axis}[
width=0.6\textwidth,
ybar,
bar width=0.15cm,
enlarge x limits=0.15,
symbolic x coords={{ETH-Small (w/ SNF)}, {ETH-Small (w/o SNF)}, {ETH-Large (w/ SNF)}, {ETH-Large (w/o SNF)}},
xtick=data,
x tick label style={rotate=30, anchor=east, font=\small},
ylabel={Illicit Class F1 Score},
ymin=0.4,
ymax=1,
legend style={at={(1.05, 0.5)}, anchor=west, legend columns=1, font=\small},
% nodes near coords,
% nodes near coords align={vertical},
% nodes near coords style={font=\tiny, rotate=90, anchor=west}, % Rotate data labels for better visibility
% grid=both, % Add grid lines
ymajorgrids=true, % Enable horizontal grid lines
grid style=dashed,
]
% Dir-GAT (max)
\addplot[fill=plotred] coordinates {
({ETH-Small (w/ SNF)}, 0.7958)
({ETH-Small (w/o SNF)}, 0.6705)
({ETH-Large (w/ SNF)}, 0.8515)
({ETH-Large (w/o SNF)}, 0.6705)
};
% Dir-GAT (cat)
\addplot[fill=plotpink] coordinates {
({ETH-Small (w/ SNF)}, 0.8831)
({ETH-Small (w/o SNF)}, 0.6705)
({ETH-Large (w/ SNF)}, 0.8769)
({ETH-Large (w/o SNF)}, 0.6705)
};

% Dir-GCN (max)
\addplot[fill=plotblue] coordinates {
({ETH-Small (w/ SNF)}, 0.8577)
({ETH-Small (w/o SNF)}, 0.9000)
({ETH-Large (w/ SNF)}, 0.8598)
({ETH-Large (w/o SNF)}, 0.9207)
};
% Dir-GCN (cat)
\addplot[fill=plotgreen] coordinates {
({ETH-Small (w/ SNF)}, 0.9240)
({ETH-Small (w/o SNF)}, 0.8987)
({ETH-Large (w/ SNF)}, 0.9168)
({ETH-Large (w/o SNF)}, 0.9188)
};

% MVGNN-\textsl{add}
\addplot[fill=plotorange] coordinates {
({ETH-Small (w/ SNF)}, 0.9231)
({ETH-Small (w/o SNF)}, 0.9333)
({ETH-Large (w/ SNF)}, 0.9300)
({ETH-Large (w/o SNF)}, 0.9365)
};
% MVGNN-\textsl{cat}
\addplot[fill=plotpurple] coordinates {
({ETH-Small (w/ SNF)}, 0.9331)
({ETH-Small (w/o SNF)}, 0.9301)
({ETH-Large (w/ SNF)}, 0.9439)
({ETH-Large (w/o SNF)}, 0.9394)
};
\legend{{Dir-GAT (max)}, {Dir-GAT (cat)}, {Dir-GCN (max)}, {Dir-GCN (cat)}, {MVGNN-add}, {MVGNN-cat}}
\end{axis}
\end{tikzpicture}
\caption{{{Illicit}} %MDPI: We have deleted the title above the image, please confirm; Please add explanation for the purple bar graph.
class F1 score of Dir-GNN family and MVGNN family models across datasets.}
\label{fig:dirgnn-family-vs-mvgnn-family}
\end{figure}

Our empirical experiments on real-world payment transaction datasets indicate that separating the two sets of learnable parameters may not always be necessary.
As shown in Figure \ref{fig:dirgnn-family-vs-mvgnn-family}, the illicit class F1 scores of our MVGNN variants consistently outperform those of the Dir-GNN variants.
On average, the F1 scores of our MVGNN-{{add}} surpasses those of the other competing methods by +7.01\%, +21.48\%, +6.22\%, and +20.75\% on the ETH-Small (w/ SNF), ETH-Small (w/o SNF), ETH-Large (w/ SNF), ETH-Large (w/o SNF), and FPT datasets, respectively.
Meanwhile, our MVGNN-{{cat}} outperforms the others by +8.17\%, +21.07\%, +7.81\%, and +21.12\% on these five datasets, respectively.

% Bar Plot: Performance of Models Across Datasets

% round 1 : reviewer 1 : point 4, lr and emb size
{\subsubsection{Effect of Learning Rate}}

\newcommand{\plotLearningRateEthSmall}{
% \begin{figure}[H]
\centering
\begin{tikzpicture}
\begin{axis}[
title={ETH-Small (w/ or w/o SNF)},
xlabel={Learning Rate},
ylabel={Illicit Class F1 Score},
xmin=0.0005, xmax=0.15,
ymin=0, ymax=1.0,
xtick={0.001, 0.01, 0.1}, % Explicit x-ticks for learning rates
legend style={legend pos=south west, legend columns=1, font=\small}, % Smaller legend font
ymajorgrids=true,
grid style=dashed,
xmode=log, % Use logarithmic scale for x-axis (learning rates)
log ticks with fixed point, % Format log ticks as regular numbers
]

% LineMVGNN-\textsl{add}, w/ SNF (False)
\addplot[
color=blue,
mark=square,
]
coordinates {
(0.001, 0.8481)(0.01, 0.9362)(0.1, 0.6141)
};
\addlegendentry{LineMVGNN-\textsl{add}, w/ SNF}

% LineMVGNN-\textsl{cat}, w/ SNF (False)
\addplot[
color=orange,
mark=triangle,
]
coordinates {
(0.001, 0.64)(0.01, 0.9325)(0.1, 0.6705)
};
\addlegendentry{LineMVGNN-\textsl{cat}, w/ SNF}

% LineMVGNN-\textsl{add}, w/o SNF (True)
\addplot[
color=green,
mark=square,
]
coordinates {
(0.001, 0.9407)(0.01, 0.8992)(0.1, 0.6705)
};
\addlegendentry{LineMVGNN-\textsl{add}, w/o SNF}

% LineMVGNN-\textsl{cat}, w/o SNF (True)
\addplot[
color=red,
mark=triangle,
]
coordinates {
(0.001, 0.8935)(0.01, 0.9222)(0.1, 0.8092)
};
\addlegendentry{LineMVGNN-\textsl{cat}, w/o SNF}

\end{axis}
\end{tikzpicture}
% \caption{Illicit class F1 against learning rate for ETH-Small (w/ or w/o SNF).}
% \label{fig:study-lr-eth-small}
% \end{figure}
}

\newcommand{\plotLearningRateEthLarge}{
% \begin{figure}[H]
\centering
\begin{tikzpicture}
\begin{axis}[
title={ETH-Large (w/ or w/o SNF)},
xlabel={Learning Rate},
ylabel={Illicit Class F1 Score},
xmin=0.0005, xmax=0.15,
ymin=0, ymax=1.0,
xtick={0.001, 0.01, 0.1}, % Explicit x-ticks for learning rates
legend style={legend pos=south west, legend columns=1, font=\small}, % Smaller legend font
ymajorgrids=true,
grid style=dashed,
xmode=log, % Use logarithmic scale for x-axis (learning rates)
log ticks with fixed point, % Format log ticks as regular numbers
]

% LineMVGNN-\textsl{add}, w/ SNF (False)
\addplot[
color=blue,
mark=square,
]
coordinates {
(0.001, 0.7975)(0.01, 0.8905)(0.1, 0.7488)
};
\addlegendentry{LineMVGNN-\textsl{add}, w/ SNF}

% LineMVGNN-\textsl{cat}, w/ SNF (False)
\addplot[
color=orange,
mark=triangle,
]
coordinates {
(0.001, 0.8588)(0.01, 0.9548)(0.1, 0.6705)
};
\addlegendentry{LineMVGNN-\textsl{cat}, w/ SNF}

% LineMVGNN-\textsl{add}, w/o SNF (True)
\addplot[
color=green,
mark=square,
]
coordinates {
(0.001, 0.7991)(0.01, 0.9309)(0.1, 0.6705)
};
\addlegendentry{LineMVGNN-\textsl{add}, w/o SNF}

% LineMVGNN-\textsl{cat}, w/o SNF (True)
\addplot[
color=red,
mark=triangle,
]
coordinates {
(0.001, 0.9402)(0.01, 0.9565)(0.1, 0.6705)
};
\addlegendentry{LineMVGNN-\textsl{cat}, w/o SNF}

\end{axis}
\end{tikzpicture}
%     \caption{Illicit class F1 against learning rate for ETH-Large (w/ or w/o SNF).}
%     \label{fig:study-lr-eth-large}
% \end{figure}
}

\newcommand{\plotLearningRateFpt}{
% \begin{figure}[H]
\centering
\begin{tikzpicture}
\begin{axis}[
title={FPT dataset},
xlabel={Learning Rate},
ylabel={Illicit Class F1 Score},
xmin=0.0007, xmax=0.15,
ymin=0., ymax=1.0,
xtick={0.001, 0.01, 0.1}, % Explicit x-ticks
legend style={legend pos=south west, legend columns=1, font=\small},
ymajorgrids=true,
grid style=dashed,
xmode=log, % Use logarithmic scale for x-axis (learning rates)
log ticks with fixed point, % Format log ticks as regular numbers
]

% LineMVGNN-\textsl{add}, w/ SNF
\addplot[
color=blue,
mark=square,
]
coordinates {
(0.001, 0.9955)(0.01, 0.9702)(0.1, 0.9314)
};
\addlegendentry{LineMVGNN-\textsl{add}}

% LineMVGNN-\textsl{cat}, w/ SNF
\addplot[
color=orange,
mark=triangle,
]
coordinates {
(0.001, 0.9954)(0.01, 0.9326)(0.1, 0.7216)
};
\addlegendentry{LineMVGNN-\textsl{cat}}

\end{axis}
\end{tikzpicture}
%     \caption{Illicit class F1 against learning rate for FPT dataset.}
%     \label{fig:study-lr-fpt}
% \end{figure}
}

\begin{figure}[H]
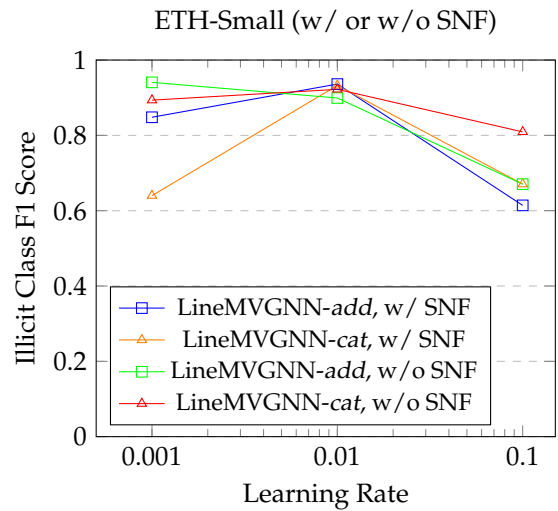

\hspace{-191pt}\plotLearningRateEthSmall % Call the macro here
\caption{Illicit class F1 against learning rate for ETH-Small (w/ or w/o SNF).}
\label{fig:study-lr-eth-small}
\end{figure}

\begin{figure}[H]
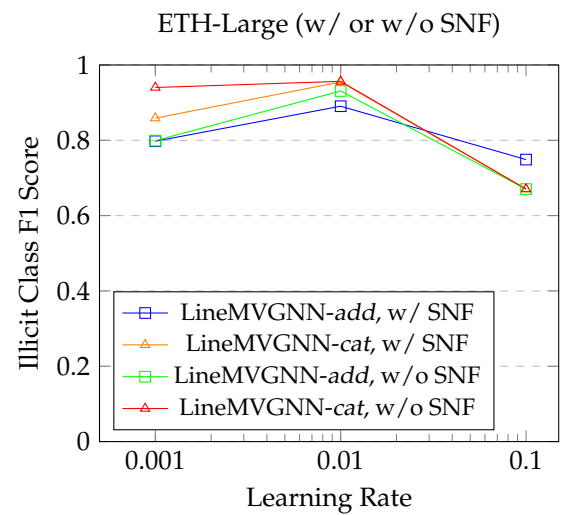

\hspace{-191pt} \plotLearningRateEthLarge % Call the macro here
\caption{Illicit class F1 against learning rate for ETH-Large (w/ or w/o SNF).}
\label{fig:study-lr-eth-large}
\end{figure}

\begin{figure}[H]
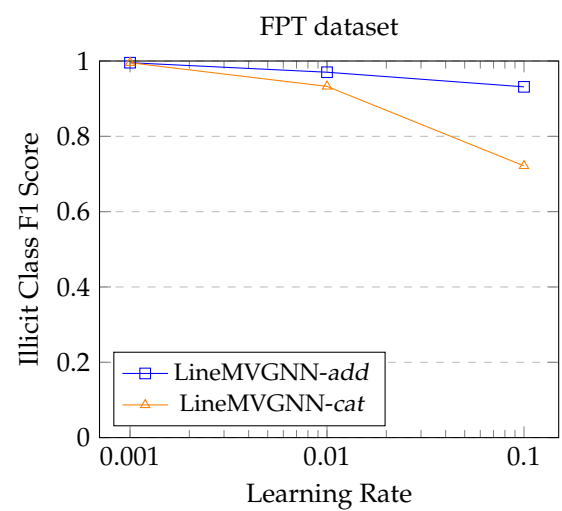

\hspace{-191pt}  \plotLearningRateFpt % Call the macro here
\caption{Illicit class F1 against learning rate for FPT.}
\label{fig:study-lr-fpt}
\end{figure}

Learning rate can influence the training dynamics and final performance of deep learning models, such as LineMVGNN-{{add}} and LineMVGNN-{{cat}}.
We conducted experiments with different learning rates chosen from ${0.1, 0.01, 0.001}$ while using default values for other hyperparameters on the ETH-Small (w/ or w/o SNF), ETH-Large (w/ or w/o SNF), and FPT datasets.
The results are summarized in Figures \ref{fig:study-lr-eth-small}--\ref{fig:study-lr-fpt}.

For the ETH-Small dataset (Figure \ref{fig:study-lr-eth-small}) and the ETH-Large dataset (Figure \ref{fig:study-lr-eth-large}), in general, the models achieve the highest F1 score for the illicit class at a learning rate of 0.01 regardless of the presence of SNF.
A large learning rate of 0.1 leads to an unstable suboptimal performance, likely due to overshooting the optimal weights during gradient descent.
Conversely, a small learning rate of 0.001 results in slower convergence and slightly lower performance in general. The models might have got trapped in local minima during training.

For the FPT dataset (Figure \ref{fig:study-lr-fpt}), the model performs best with a small learning rate of 0.001. This dataset is more complex, with more attributes and a larger number of nodes and edges. A smaller learning rate allows for more stable training, reducing the risk of overshooting the optimal solution, leading to better generalization. A large learning rate of 0.1 results in poor performance due to the instability of training, while a learning rate of 0.01 performs well but not as effectively as 0.001.

Comparing LineMVGNN-{{cat}} and LineMVGNN-{{add}}, although both variants exhibit similar behavior under the variation of learning rates, LineMVGNN-{{add}} is more robust to the variation of learning rate, especially when the transaction graph data is more complex.
As shown in Figure \ref{fig:study-lr-fpt} for the FPT dataset, although both variant models score over 0.99 for the illicit class F1, LineMVGNN-{{add}}'s performance degrades less noticeably than LineMVGNN-{{cat}} when the learning rate increases from 0.001 to 0.1.
This reflects that LineMVGNN-{{cat}}, which combines messages from in- and out-neighbors via concatenation followed by a linear transformation, requires more careful tuning to ensure stable convergence.
The linear transformation layer in LineMVGNN-{{cat}} introduces more learnable parameters, making the model more susceptible to oscillations or divergence if the learning rate is too high.

\vspace{-6pt}

\vspace{-6pt}

\subsubsection{Effect of Embedding Size}

The embedding size determines the dimensionality of the node and edge representations learned by the models.
Generally, a larger embedding size can capture more complex patterns and nuances in the data.
To explore the impact of embedding size on the performance of our LineMVGNN variants, we conduct experiments with different embedding sizes on the FPT dataset.

\begin{figure}[H]

\hspace{-4pt}  \begin{tikzpicture}
\begin{axis}[
xlabel={Embedding Size},
ylabel={Illicit Class F1 Score},
xmin=13, xmax=80,   % xmin=10, xmax=70,
ymin=0.95, ymax=1.0,
xtick={16, 32, 64}, % Explicit x-ticks
ytick distance=0.01, % Auto y-ticks with spacing of 0.01
legend style={legend pos=south west, legend columns=1, font=\small},
% legend font=\small
ymajorgrids=true,
grid style=dashed,
xmode=log, % Use logarithmic scale for x-axis (learning rates)
log ticks with fixed point, % Format log ticks as regular numbers
]

% LineMVGNN-\textsl{add}, w/ SNF
\addplot[
color=blue,
mark=square,
]
coordinates {
(16.0, 0.9736)(32.0, 0.9948)(64.0, 0.9955)
};
\addlegendentry{LineMVGNN-\textsl{add}}

% LineMVGNN-\textsl{cat}, w/ SNF
\addplot[
color=orange,
mark=triangle,
]
coordinates {
(16.0, 0.984)(32.0, 0.9937)(64.0, 0.9954)
};
\addlegendentry{LineMVGNN-\textsl{cat}}

\end{axis}
\end{tikzpicture}
\caption{{{Illicit}} %MDPI: We have deleted the title above the image, please confirm.
class F1 against embedding size for FPT dataset (w/ SNF).}
\label{fig:study-emb-size-fpt}
\end{figure}
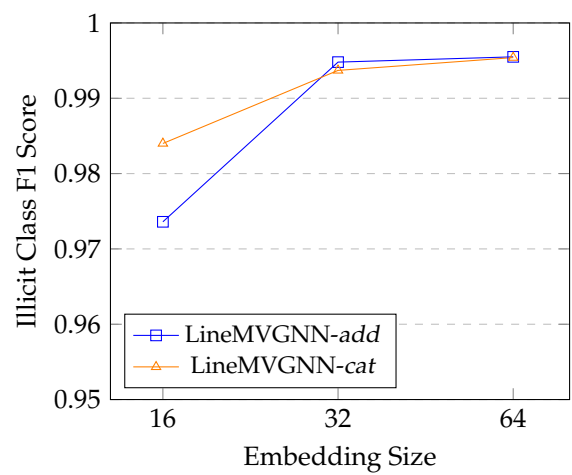

As shown in Figure \ref{fig:study-emb-size-fpt}, we vary embedding sizes of 16, 32, and 64.
The models achieve the highest F1 score for the illicit class with an embedding size of 64.
An embedding size of 32 results in a significant drop in the F1 score, likely due to insufficient capacity to capture the complex relationships in the transaction graph.
On the other hand, doubling the embedding size from 32 to 64 provides little improvement: $+0.1711\%$ and $+0.0704\%$ for LineMVGNN-{{cat}} and LineMVGNN-{{add}}, respectively.
The results suggest that an embedding size of 32 or 64 provides a good balance between model capacity and computational efficiency for the FPT dataset, while avoiding the pitfalls of overfitting and excessive computational cost associated with larger embedding sizes.

Comparing LineMVGNN-{{cat}} and LineMVGNN-{{add}}, LineMVGNN-{{cat}}, which leverages concatenation for message combination, is less sensitive to the increase in embedding size when the size doubles from 16.
In contrast, LineMVGNN-{{add}}, which uses a weighted sum for message aggregation, is slightly less robustness. Its performance drops more noticeably with smaller embedding sizes.
This suggests that the weighted sum mechanism may struggle to capture sufficient information with lower-dimensional representations.
This difference highlights the advantage of LineMVGNN-{{cat}}'s concatenation-based approach, which provides greater flexibility in combining features and is less sensitive to embedding size variations.

Overall, both models benefit from an embedding size of 64, but LineMVGNN-{{cat}} demonstrates superior stability and performance across different embedding sizes, particularly in more complex datasets like FPT.
We did not experiment on the ETH-Small and ETH-Large datasets due to the lack of attributes in the datasets.
There are only three attributes in the datasets, which are transaction timestamps and amounts, and a label to indicate all fraud nodes.
Given the limited feature set, the embedding size was set to match the dimensionality of the available attributes, ensuring that the model could effectively capture the sparse information present in the ETH datasets.

% \begin{figure}[H]
%     \centering
%     \includegraphics[width=0.7 \linewidth]{figs/study-emb-size-fpt-v2.png}
%     \caption{Illicit class F1 against embedding size for FPT dataset.}
%    \label{fig:study-emb-size-fpt}
% \end{figure}

% round 1: reviewer 1: qualitative discussion
\subsubsection{{Qualitative Discussion}}
{
LineMVGNN improves over existing methods primarily through its ability to effectively propagate and leverage transaction-level information, which is crucial for AML detection.
Unlike traditional GNNs that focus solely on node-level interactions, LineMVGNN introduces a line graph view that explicitly models edge (transaction) adjacencies.
This allows the model to capture the flow of money between transactions, which is essential for identifying suspicious patterns such as temporary repositories of funds or cyclic transactions.
By propagating edge features before updating node features, LineMVGNN ensures that transaction-level details are preserved and utilized in the learning process.
Additionally, the two-way message passing mechanism in LineMVGNN (and in MVGNN), which considers both in- and out-neighbors, enhances the model's ability to capture the directional flow of funds, further improving its detection capabilities.
These features collectively enable LineMVGNN to outperform other competing methods, as demonstrated by its superior performance on the ETH-Small (w/ or w/o SNF), ETH-Large (w/ or w/o SNF), and FPT datasets.
}

%  round 1 : reviewer 1 : point 3 and 5, about scalability, adversarial robustness, and regulatory considerations (explainability?)

\section{Limitations and Future Work}
\label{sec5}
\subsection{Scalability}

While the asymptotic computational complexity of our refined LineMVGNN model has the same asymptotic complexity $\mathcal{O}(L|\mathcal{E}|)$ as GCN \cite{Kipf:2016tc},
the refined LineMVGNN may have a higher practical runtime due to the additional overhead of edge propagation.
Processing the entire graph in a single batch may not be feasible for extremely large graphs due to memory constraints.
To alleviate this, graph sampling techniques can be leveraged, yielding smaller subgraphs for minibatch training.

\subsection{Adversarial Robustness}

In financial applications such as fraud detection, malicious actors may attempt to manipulate the graph to evade detection.
Our current setup does not apply adversarial training for the model.
In our future work, we can generate perturbed versions of the graphs during training and optimizing the model to perform well on both clean and perturbed data.
On the other hand, before applying the model, graph purification can be integrated into the model, detecting and removing adversarial edges (transactions) or nodes (accounts).
These techniques can enhance the model's resilience to adversarial~perturbations.

\subsection{Regulatory Considerations}

When deploying neural network models in highly regulated domains such as finance, the transparency and traceability of decisions are paramount.
The proposed LineMVGNN model inherently provides explainability by explicitly modeling and propagating edge-level information (e.g., transactions) and aggregating it to the node level (e.g., accounts).
This design allows the model to propagate and compare transactions (edge information) related to the same account (node). This highlights the contributions of individual edges (transactions) to the final node representations, making its decision-making process interpretable.
For further improvement, explainable artificial intelligence (XAI) techniques can be integrated into the model, such as attention mechanisms and post hoc explainability~methods.

\section{Conclusions}
\label{sec6}
Existing GNNs for digraphs face different challenges such as ignoring multi-dimensional edge features, the lack of model interpretability, and suboptimal model efficiency.
To address these issues, we introduce a lighter-weight yet effective model, {{MVGNN}}, which shares parameters in the aggregation maps for payment and receipt transactions.
This effectively and efficiently leverages the local original view and the local reversed view.
Extended from MVGNN, we propose {{LineMVGNN}} which leverages line graph transformation for the enhanced propagation of transaction information.
LineMVGNN surpasses SOTA methods in detecting money laundering activities and other financial frauds in real-world datasets.
% Potential directions for future work include exploring other message combination functions and studying the effect of graph sampling on the model performance.

%%%%%%%%%%%%%%%%%%%%%%%%%%%%%%%%%%%%%%%%%%
\vspace{6pt}

%%%%%%%%%%%%%%%%%%%%%%%%%%%%%%%%%%%%%%%%%%
% % optional
% \supplementary{The following supporting information can be downloaded at:  \linksupplementary{s1}, Figure S1: title; Table S1: title; Video S1: title.}


% Only used for preprtints:
% \supplementary{The following supporting information can be downloaded at the website of this paper posted on \href{https://www.preprints.org/}{Preprints.org}.}

% \begin{tabularx}{\textwidth}{lll}
% \toprule
% \textbf{Name} & \textbf{Type} & \textbf{Description} \\
% \midrule
% S1 & Python script (.py) & Script of python source code used in XX \\
% S2 & Text (.txt) & Script of modelling code used to make Figure X \\
% S3 & Text (.txt) & Raw data from experiment X \\
% S4 & Video (.mp4) & Video demonstrating the hardware in use \\
% ... & ... & ... \\
% \bottomrule
% \end{tabularx}
% }

%%%%%%%%%%%%%%%%%%%%%%%%%%%%%%%%%%%%%%%%%%
\authorcontributions{Data curation, C.-H.P. and J.-H.C.;
Funding acquisition, C.C.;
Investigation, C.-H.P. and J.-H.C.;
Methodology, C.-H.P.;
Project administration, C.C.;
Software, C.-H.P. and J.-H.C.;
Supervision, J.K. and C.C.;
Writing---original draft, C.-H.P.;
Writing---review and editing, C.-H.P. and J.K. All authors have read and agreed to the published version of the manuscript.}

% (ITP/050/22LP)
\funding{{{This}} %MDPI: Information regarding the funder and the funding number should be provided. Please check the accuracy of funding data and any other information carefully.
%author: after consulting LSCM supervisors, the funding number cannot be provided.
research was funded by the Innovation and Technology Fund of the Hong Kong Special Administrative Region. The article processing charges were funded by Logistics and Supply Chain MultiTech R\&D Centre.}

\institutionalreview{Not applicable.}

\informedconsent{Not applicable.}

\dataavailability{The Ethereum Phishing Transaction Network (ETH) dataset is available {at} %MDPI: Please add the access date (Format: Date Month Year). e.g., (accessed on 1 January 2020).
\url{https://www.kaggle.com/datasets/xblock/ethereum-phishing-transaction-network} {({accessed on 1 January 2025)}}.
The Financial Payment Transaction (FPT) Dataset is not readily available because of privacy regulations.}

% \acknowledgments{{This} %MDPI: Please ensure that all individuals included in this section have consented to the acknowledgement; {} %MDPI: To AE: please confirm if the funding information in the Acknowledgments Section should be moved to the Funding Section.
%  research is supported by the Logistics and Supply Chain MultiTech R\&D~Centre.
% }

\acknowledgments{
%MDPI: Please ensure that all individuals included in this section have consented to the acknowledgement; {} %MDPI: To AE: please confirm if the funding information in the Acknowledgments Section should be moved to the Funding Section.
We gratefully acknowledge the Logistics and Supply Chain MultiTech R\&D Centre for providing computational resources and supporting the article processing charges for this~publication.
}

\conflictsofinterest{The authors declare no conflicts of interest.}

%%%%%%%%%%%%%%%%%%%%%%%%%%%%%%%%%%%%%%%%%%
%% Optional

%% Only for journal Encyclopedia
%\entrylink{The Link to this entry published on the encyclopedia platform.}

\abbreviations{Abbreviations}{
The following abbreviations are used in this manuscript:\\

\noindent
\begin{tabular}{@{}ll}
SOTA & State-of-the-Art\\
AML & Anti-Money Laundering\\
GNN & Graph Neural Network\\
ETH & Ethereum\\
FPT & Financial Payment Transaction\\
SNF & Structural Node Feature\\
MPNN & Message Passing Neural Network\\
Dir-GNN & Directed Graph Neural Network\\
MVGNN & Multi-View Graph Neural Network\\
LineMVGNN & Line-Graph-Assisted Multi-View Graph Neural Network\\
GCN & Graph Convolutional Network\\
GraphSAGE & Graph SAmple and aggreGatE\\
GIN & Graph Isomorphism Network\\
PNA & Principal Neighborhood Aggregation\\
EGAT & Graph Attention Network with Edge Features\\
DiGCN & Digraph Inception Convolutional Networks\\
Dir-GCN & Directed Graph Convolutional Network\\
Dir-GAT & Directed Graph Attention Network\\
MagNet & Digraph GNN Based on the Magnetic Laplacian\\
SigMaNet & Digraph GNN Based on the Sign-Magnetic Laplacian\\
FaberNet & Spectral Digraph GNN Using Faber Polynomials\\
OOM & Out of Memory\\
TWMP & Two-Way Message Passing\\
LGV & Line Graph View\\
eq & equation\\
w/ & with\\
w/o & without
\end{tabular}
}

%%%%%%%%%%%%%%%%%%%%%%%%%%%%%%%%%%%%%%%%%%
%% Optional
\appendixtitles{yes} % Leave argument "no" if all appendix headings stay EMPTY (then no dot is printed after "Appendix A"). If the appendix sections contain a heading then change the argument to "yes".
\appendixstart
\appendix

\section[\appendixname~\thesection]{FPT Dataset}
\label{sec: FPT Dataset}

{{There}} %MDPI: Please add the citation of the appendix in the main text.
are a total of 43 attributes in the dataset. Fourteen relevant attributes are extracted to construct a transaction graph for each day: currency, transaction type, business service, payment category purpose, status, reject reason code, return reason code, outward input source, inward delivery channel, real-time counterparty verification, settlement amount, settlement time, credit participant account type, and debit participant account~type.

Table \ref{table: fpt dataset stats} shows detailed statistics of each graph of each day in our FPT dataset.

\begin{table}[H]
\setlength{\tabcolsep}{8.3mm}    %{11.6mm}
\centering
\caption{Graph statistics of the FPT dataset.}
\small
\begin{tabular}{cccc}
\toprule
\textbf{Set} & \textbf{Day} & \textbf{Number of Nodes} & \textbf{Number of Edges} \\

\midrule
\multirow{19}{*}{Train}
& 1 & {{1,116,969}} %MDPI: Please confirm whether the numbers with more than five digits in the last two columns need to be separated by commas.
& {{1,160,635}} \\
& 2 & 1,013,391 & 1,039,540 \\
& 3 & 1,259,733 & 1,294,309 \\
& 4 & 1,175,766 & 1,208,983 \\
& 5 & 1,165,737 & 1,217,110 \\
& 6 & 1,101,062 & 1,141,048 \\
& 7 & 1,137,598 & 1,185,835 \\
& 8 & 911,029 & 955,756 \\
& 9 & 924,847 & 976,963 \\
& 10 & 1,117,958 & 1,167,333 \\
& 11 & 997,538 & 1,037,538 \\
& 12 & 1,036,556 & 1,094,068 \\
& 13 & 970,965 & 1,008,168 \\
& 14 & 976,630 & 1,012,067 \\
& 15 & 888,321 & 920,889 \\
& 16 & 875,318 & 925,757 \\
& 17 & 1,029,538 & 1,070,001 \\
& 18 & 975,762 & 1,012,953 \\
& 19 & 1,024,562 & 1,077,209 \\
\midrule
\multirow{6}{*}{Validation}
& 20 & 1,024,570 & 1,061,908 \\
& 21 & 982,044 & 1,025,592 \\
& 22 & 843,878 & 879,405 \\
& 23 & 848,317 & 902,103 \\
& 24 & 1,044,676 & 1,094,133 \\
& 25 & 1,057,020 & 1,097,454 \\
\midrule
\multirow{6}{*}{Test}
& 26 & 1,117,969 & 1,176,023 \\
& 27 & 1,173,160 & 1,221,627 \\
& 28 & 1,265,930 & 1,314,902 \\
& 29 & 1,022,037 & 1,064,851 \\
& 30 & 1,001,363 & 1,060,849 \\
& 31 & 1,423,624 & 1,474,741 \\
\bottomrule
\end{tabular}
\label{table: fpt dataset stats}
\end{table}

\begin{figure}[H]    %[!t]
% \centering
\includegraphics[width=0.5\linewidth]{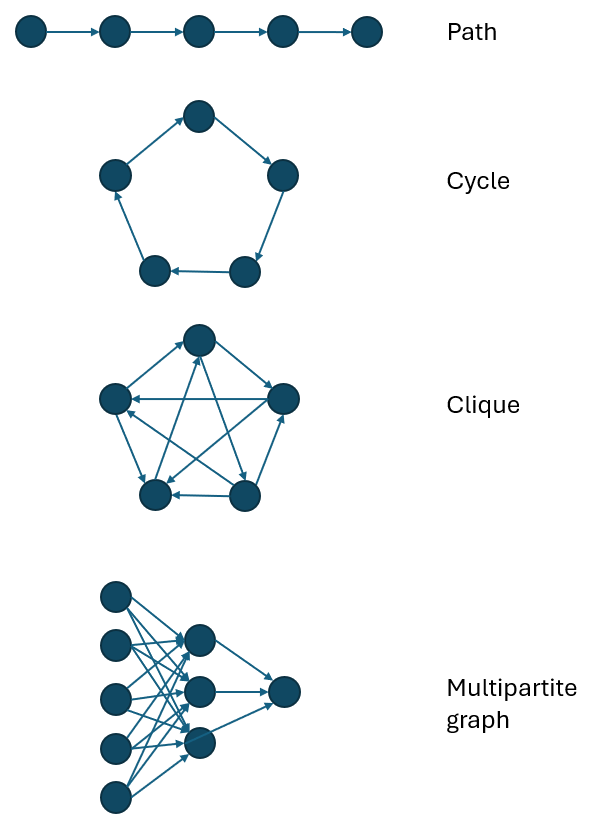}
\caption{Embedded money laundering patterns following \cite{Elliott2019AnomalyDI}.}
\label{fig:embedded-synthetic-patterns}
\end{figure}

As we assume that the real data contain no anomalous money laundering patterns, synthetic anomalies are injected into the graphs. Following the anomaly injection strategy from \cite{Elliott2019AnomalyDI}, we embed directed paths, cycles (rings), cliques, and multipartite structures as shown in Figure \ref{fig:embedded-synthetic-patterns}. The sizes of paths and cycles are randomly chosen from 10 to 20, while the size of cliques ranges from 5 to 10. The source, intermediate, and destination layer in a directed multipartite network have fixed sizes of 5, 3, and 1 respectively. The steps of anomaly synthesis for each transaction graph of a day are as follows:

\begin{enumerate}
\item Randomly select a pattern from {path, cycle, clique, multipartite graph}.
\item Generate the pattern with n nodes (and e edges).
\item Randomly select e rows of transaction data from the FPT dataset.
\item Assign each row of transaction attribute values to each edge and the corresponding end nodes. {{(To simulate a flow of money in paths and cycles, the selected rows of transaction data are sorted and assigned, such that for each node the transaction timestamp of incoming edges is earlier than that of the outgoing edges except one edge in each cycle pattern. Similarly, in multipartite graphs, the selected transaction data are sorted and assigned such that transaction timestamps in all edges in the first layer are earlier than those in the second layer. Also, in each path and cycle, transaction amounts of edges within a given anomaly are set by randomly choosing from one out of e rows of selected transaction data).}} %MDPI: The journal does not allow the note format, so we moved it to the main text, please confirm.
\item Insert the anomaly into the transaction graph.
\item Repeat the steps above until a desired number of synthetic nodes has been reached.
\end{enumerate}
\vspace{-9pt}

\section[\appendixname~\thesubsection]{Compared Methods}
\label{sec: appendix-compared-methods}
\begin{itemize}

\item {{GCN}} \cite{Kipf:2016tc} leverages spectral graph convolutions to capture neighborhood information and perform various tasks, such as node classification.
% We also create a variant \textsl{{GCN (no norm.)}} which is a GCN without adjacency normalization.
Since it does not naturally support multi-dimensional edge features, we concatenate in-node features with in-edge features during message passing.

\item {{GraphSAGE}} \cite{HamiltonYL17} utilizes neighborhood sampling and aggregation for inductive learning on large graphs. We choose the pooling aggregator and full neighbor sampling as the baseline model setting. Since it does not naturally support multi-dimensional edge features, we concatenate in-node features with in-edge features during \mbox{message~passing. }

\item {{MPNN}} \cite{gilmer2017neural} is a framework for processing graph structured data. It enables the exchange of messages between nodes iteratively, allowing for information aggregation and updates. Specifically, it proposes the message function to be a matrix multiplication between the source node embeddings and a matrix, which is mapped by the edge feature vectors with a neural network.

% \item \textbf{SGC} \cite{pmlr-v97-wu19e} [ === WIP === ] [Skip]

\item {{GIN}} \cite{journals/corr/abs-1810-00826} is designed to achieve maximum discriminative power among WL-test equivalent graphs. It uses sum aggregation and MLPs to process node features and neighborhood information. Since it does not naturally support multi-dimensional edge features, we concatenate in-node features with in-edge features during message passing. Although \cite{HuLGZLPL20} extends GIN by summing up node features and edge features, we find it inappropriate for our datasets because of (1) the considerable context difference between node and edge features and (2) the difference in feature sizes.

\item {{PNA}} \cite{CorsoCBLV20} enhances GNNs by employing multiple aggregators and degree scalars. For aggregators, we picked mean, max, min, and sum; for degree scalars, amplification, attenuation, and identity are used.

\item {{EGAT}} \cite{10.1093/bib/bbab371} extends graph attention networks, GAT, by incorporating edge features into the attention mechanism. The unnormalized attention score is computed with a concatenated vector of node and edge features. In this work, we use three attention heads by default.

\item {{DiGCN}} \cite{conf/nips/TongLSLR020} extends graph convolutional networks to digraphs. It utilizes digraph convolution and kth-order proximity to achieve larger receptive fields and learn multi-scale features in digraphs. As suggested in the paper, we compute the approximate digraph Laplacian, which alters node connections, during data preprocessing because of considerable computation time. Since it does not naturally support multi-dimensional edge features and performs edge manipulation (such as adding/removing edges), we aggregate all features from in-edges by summation and update node features by concatenating with the aggregated edge features.

\item {{MagNet}} \cite{conf/nips/ZhangHBPH21} is a spectral GNN for digraphs that utilizes a complex Hermitian matrix called the magnetic Laplacian to encode both undirected structure and directional information. We set the phase parameter $q = [0, 0.25]$ to be learnable and initialize it as 0.125. Unless otherwise specified, other model parameters are set to default values from
{{PyTorch Geometric Signed Directed (version 0.22.0)}} %MDPI: Please state the software version number.
\cite{PyTorchGeometricSignedDirected2023}. Since it does not naturally support multi-dimensional edge features and performs edge manipulation (such as adding/removing edges), we aggregate all features from in-edges by summation and update node features by concatenating with the aggregated edge features.

\item {{SigMaNet}} \cite{10.1609/aaai.v37i6.25919} is a generalized graph convolutional network that unifies the treatment of undirected and directed graphs with arbitrary edge weights. It introduces the Sign-Magnetic Laplacian which extends spectral GCN theory to graphs with positive and negative weights. Since it does not naturally support multi-dimensional edge features and performs edge manipulation (such as adding/removing edges), we aggregate all features from in-edges by summation and update node features by concatenating with the aggregated edge features.

\item {{FaberNet}} \cite{journals/corr/abs-2310-02232} leverages Faber Polynomials and advanced tools from complex analysis to extend spectral convolutional networks to digraphs. It achieves superior results in heterophilic node classification. Unless specified, default parameters in that paper are used. We experimented with two different jumping knowledge options (``cat'' and ``max''), producing variant models {{FaberNet (cat)}} and {{FaberNet (cat)}}, respectively. We use real FaberNets because \cite{journals/corr/abs-2310-02232} proves that the expressive power of real FaberNets is higher than complex ones given the same number of real parameters. Since it does not naturally support multi-dimensional edge features, we concatenate in-node features with in-edge features during message passing.

\item {{Dir-GCN}} and \textsl{Dir-GAT} \cite{conf/log/RossiCGFGB23} are instance models under the proposed Dir-GNN framework for digraph learning. It extends message passing neural networks by performing separate message aggregations from in- and out-neighbors. We experimented on the base models, GCN and GAT, respectively, with two different jumping knowledge options (``max'' and ``cat'') with learnable combination coefficient $\alpha$, producing four variant models, namely {{Dir-GCN (cat)}}, {{Dir-GCN (max)}}, {{Dir-GAT (max)}}, and {{Dir-GAT (cat)}}. For details about jumping knowledge, readers can refer to \cite{Xu:2018vn}.

\end{itemize}

\section[\appendixname~\thesubsection]{Implementation Details}
\label{sec: appendix-implementation-details}
PyTorch \cite{paszke2019pytorch}, Deep Graph Library (DGL) \cite{wang2020deep}, and PyTorch Geometric (PyG) \cite{fey2019graph} are used for implementing all algorithms on a single NVIDIA GV100GL [Tesla V100 SXM3 32~GB] GPU. For all GNN models, the depth is 2 by default. We train the models with the {{Adam}} optimizer \cite{KingBa15}.
We use cosine annealing with warm restarts at epoch 10, 20, 40, and so on~\cite{conf/iclr/LoshchilovH17}.
Cross-entropy loss is used.
The initial learning rate, $lr$, is optimized by grid search, where $lr \in \{0.1, 0.01, 0.001\}$.

For the FPT dataset, the maximum number of epochs is 500 with an early stopping patience of 25 epochs on the validation loss. Due to class imbalance, weighted cross-entropy loss is used where the weight for each class is the inverse of the square root of the number of samples belonging to that class. By default, the node embedding size is 64.

For the Ethereum datasets, the maximum number of epochs is 5000 with an early stopping patience of 500 epochs on the validation loss. The node embedding size is the same as the edge feature dimension. Due to the huge number of edges in line graphs, edge sampling is adopted for nodes whose in-degrees exceed $\tau$, where $\tau$ is optimized by grid search. For Eth-Small, $\tau \in \{100, 200, 500, 1000\}$; for Eth-Large, $\tau \in \{500, 1000, 2000, 5000\}$.

%%%%%%%%%%%%%%%%%%%%%%%%%%%%%%%%%%%%%%%%%%
%\isPreprints{}{% This command is only used for ``preprints''.
\begin{adjustwidth}{-\extralength}{0cm}

\reftitle{References}



% If authors have biography, please use the format below
%\section*{Short Biography of Authors}
%\bio
%{\raisebox{-0.35cm}{\includegraphics[width=3.5cm,height=5.3cm,clip,keepaspectratio]{Definitions/author1.pdf}}}
%{\textbf{Firstname Lastname} Biography of first author}
%
%\bio
%{\raisebox{-0.35cm}{\includegraphics[width=3.5cm,height=5.3cm,clip,keepaspectratio]{Definitions/author2.jpg}}}
%{\textbf{Firstname Lastname} Biography of second author}

% For the MDPI journals use author-date citation, please follow the formatting guidelines on http://www.mdpi.com/authors/references
% To cite two works by the same author: \citeauthor{ref-journal-1a} (\citeyear{ref-journal-1a}, \citeyear{ref-journal-1b}). This produces: Whittaker (1967, 1975)
% To cite two works by the same author with specific pages: \citeauthor{ref-journal-3a} (\citeyear{ref-journal-3a}, p. 328; \citeyear{ref-journal-3b}, p.475). This produces: Wong (1999, p. 328; 2000, p. 475)

%
%
%\reviewreports{\\
%Reviewer 1 comments and authors’ response\\
%Reviewer 2 comments and authors’ response\\
%Reviewer 3 comments and authors’ response
%}
%
\PublishersNote{}
%\isPreprints{}{
\end{adjustwidth}
%} 

\begin{thebibliography}{999}

\bibitem[Chen et~al.(2018)Chen, Khoa, Teoh, Nazir, Karuppiah, and
Lam]{10.1007/s10115-017-1144-z}
Chen, Z.; Khoa, L.D.; Teoh, E.N.; Nazir, A.; Karuppiah, E.K.; Lam, K.S.
\newblock Machine learning techniques for anti-money laundering (AML) solutions
in suspicious transaction detection: A review.
\newblock {\em Knowl. Inf. Syst.} {\bf 2018}, {\em 57},~245--285. [\href{http://doi.org/10.1007/s10115-017-1144-z}{CrossRef}]

\bibitem[Hilal et~al.(2022)Hilal, Gadsden, and Yawney]{HILAL2022116429}
Hilal, W.; Gadsden, S.A.; Yawney, J.
\newblock Financial Fraud: A Review of Anomaly Detection Techniques and Recent
Advances.
\newblock {\em Expert Syst. Appl.} {\bf 2022}, {\em 193},~116429. [\href{http://dx.doi.org/10.1016/j.eswa.2021.116429}{CrossRef}]

\bibitem[Motie and Raahemi(2024)]{MOTIE2024122156}
Motie, S.; Raahemi, B.
\newblock Financial fraud detection using graph neural networks: {A} systematic
review.
\newblock {\em Expert Syst. Appl.} {\bf 2024}, {\em 240},~122156. [\href{http://dx.doi.org/10.1016/J.ESWA.2023.122156}{CrossRef}]

\bibitem[Koke and Cremers(2024)]{journals/corr/abs-2310-02232}
Koke, C.; Cremers, D.
\newblock HoloNets: Spectral Convolutions do extend to Directed Graphs.
\newblock In Proceedings of the  Twelfth International Conference on
Learning Representations, {ICLR} 2024, Vienna, Austria,  7--11 May 2024.


\bibitem[Fiorini et~al.(2023)Fiorini, Coniglio, Ciavotta, and
Messina]{10.1609/aaai.v37i6.25919}
Fiorini, S.; Coniglio, S.; Ciavotta, M.; Messina, E.
\newblock SigMaNet: One laplacian to rule them all.
\newblock In Proceedings of the Thirty-Seventh AAAI
Conference on Artificial Intelligence and Thirty-Fifth Conference on
Innovative Applications of Artificial Intelligence and Thirteenth Symposium
on Educational Advances in Artificial Intelligence,  AAAI'23/IAAI'23/EAAI'23, {{ Washington, DC, USA, 7--14 February 2023}};  AAAI Press: {{Washington, DC, USA}}, 2023. [\href{http://dx.doi.org/10.1609/aaai.v37i6.25919}{CrossRef}]

\bibitem[Zhang et~al.(2021)Zhang, He, Brugnone, Perlmutter, and
Hirn]{conf/nips/ZhangHBPH21}
Zhang, X.; He, Y.; Brugnone, N.; Perlmutter, M.; Hirn, M.J.
\newblock MagNet: A Neural Network for Directed Graphs.
\newblock In Proceedings of the NeurIPS, {{Online,  6--14 December}} 2021; Ranzato, M., Beygelzimer, A., Dauphin, Y.N., Liang, P., Vaughan, J.W., Eds.;
{{Curran Associates, Inc.: Red Hook, NY, USA, 2021}}; \mbox{pp. 27003--27015.}

\bibitem[Tong et~al.(2020{\natexlab{a}})Tong, Liang, Sun, Li, Rosenblum, and
Lim]{conf/nips/TongLSLR020}
Tong, Z.; Liang, Y.; Sun, C.; Li, X.; Rosenblum, D.S.; Lim, A.
\newblock Digraph Inception Convolutional Networks.
\newblock In Proceedings of the Advances in Neural Information Processing
Systems 33: Annual Conference on Neural Information Processing Systems 2020,
NeurIPS 2020, Virtual, 6--12 December 2020; Larochelle, H., Ranzato, M.,
Hadsell, R., Balcan, M., Lin, H., Eds.;
{{Curran Associates, Inc.: Red Hook, NY, USA, 2020}}.

\bibitem[Tong et~al.(2020{\natexlab{b}})Tong, Liang, Sun, Rosenblum, and
Lim]{journals/corr/abs-2004-13970}
Tong, Z.; Liang, Y.; Sun, C.; Rosenblum, D.S.; Lim, A.
\newblock Directed Graph Convolutional Network.
\newblock {\em arXiv} {\bf 2020},  arXiv:2004.13970.


\bibitem[Ma et~al.(2019)Ma, Hao, Yang, Li, Jin, and
Chen]{Ma2019SpectralbasedGC}
Ma, Y.; Hao, J.; Yang, Y.; Li, H.; Jin, J.; Chen, G.
\newblock Spectral-based Graph Convolutional Network for Directed Graphs.
\newblock {\em arXiv} {\bf 2019}, arXiv:1907.08990.


\bibitem[Monti et~al.(2018)Monti, Otness, and
Bronstein]{DBLP:journals/corr/abs-1802-01572}
Monti, F.; Otness, K.; Bronstein, M.M.
\newblock MotifNet: A motif-based Graph Convolutional Network for directed
graphs.
\newblock {\em arXiv} {\bf 2018}, arXiv:1802.01572.


\bibitem[Rossi et~al.(2023)Rossi, Charpentier, Giovanni, Frasca, Günnemann,
and Bronstein]{conf/log/RossiCGFGB23}
Rossi, E.; Charpentier, B.; Giovanni, F.D.; Frasca, F.; Günnemann, S.;
Bronstein, M.M.
\newblock Edge Directionality Improves Learning on Heterophilic Graphs.
\newblock In Proceedings of the LoG, PMLR, {{Virtual,  27--30 November 2023}}; Villar, S., Chamberlain, B., Eds.; {{PMLR: London, UK, 2023}}; Volume 231, p.~25.

\bibitem[Thost and Chen(2021)]{conf/iclr/Thost021}
Thost, V.; Chen, J.
\newblock Directed Acyclic Graph Neural Networks.
\newblock In Proceedings of the 9th International Conference on Learning
Representations, {ICLR} 2021, Virtual, Austria,  3--7 May 2021.


\bibitem[Li et~al.(2016)Li, Tarlow, Brockschmidt, and
Zemel]{journals/corr/LiTBZ15}
Li, Y.; Tarlow, D.; Brockschmidt, M.; Zemel, R.S.
\newblock Gated Graph Sequence Neural Networks.
\newblock In Proceedings of the 4th International Conference on Learning
Representations, {ICLR} 2016, San Juan, PR, USA,  2--4 May 2016.
% Conference Track Proceedings; Bengio, Y., LeCun, Y., Eds.;
% {{ICLR, San Juan, PR, USA, 2--4 May 2016}} 


\bibitem[Gilmer et~al.(2017)Gilmer, Schoenholz, Riley, Vinyals, and
Dahl]{gilmer2017neural}
Gilmer, J.; Schoenholz, S.S.; Riley, P.F.; Vinyals, O.; Dahl, G.E.
\newblock Neural Message Passing for Quantum Chemistry.
\newblock In Proceedings of the 34th International
Conference on Machine Learning, {ICML} 2017, Sydney, NSW, Australia, 6--11
August 2017; Precup, D., Teh, Y.W., Eds.; PMLR: {{London, UK}}, 2017; Volume 70, pp. 1263--1272.

\bibitem[Scarselli et~al.(2009)Scarselli, Gori, Tsoi, Hagenbuchner, and
Monfardini]{4700287}
Scarselli, F.; Gori, M.; Tsoi, A.C.; Hagenbuchner, M.; Monfardini, G.
\newblock The Graph Neural Network Model.
\newblock {\em IEEE Trans. Neural Netw.} {\bf 2009}, {\em
20},~61--80. [\href{http://dx.doi.org/10.1109/TNN.2008.2005605}{CrossRef}] [\href{http://www.ncbi.nlm.nih.gov/pubmed/19068426}{PubMed}]

\bibitem[{Joint Financial Intelligence Unit}(2024)]{jfiu2024}
{Joint Financial Intelligence Unit}.
\newblock Screen the Account for Suspicious Indicators: Recognition of a
Suspicious Activity Indicator or Indicators.  2024.
Available online: \url{https://www.jfiu.gov.hk/en/str\_screen.html} (accessed on 10 August 2024).

\bibitem[Bahmani et~al.(2010)Bahmani, Chowdhury, and
Goel]{10.14778/1929861.1929864}
Bahmani, B.; Chowdhury, A.; Goel, A.
\newblock Fast incremental and personalized PageRank.
\newblock {\em Proc. VLDB Endow.} {\bf 2010}, {\em 4},~173--184. [\href{http://dx.doi.org/10.14778/1929861.1929864}{CrossRef}]

\bibitem[Velickovic et~al.(2018)Velickovic, Cucurull, Casanova, Romero,
Li{\`{o}}, and Bengio]{journals/corr/abs-1710-10903}
Velickovic, P.; Cucurull, G.; Casanova, A.; Romero, A.; Li{\`{o}}, P.; Bengio,
Y.
\newblock Graph Attention Networks.
\newblock In Proceedings of the 6th International Conference on Learning
Representations, {ICLR} 2018, Vancouver, BC, Canada,  30 April--3 May 2018.

\bibitem[Chen et~al.(2019)Chen, Li, and Bruna]{chen2017supervised}
Chen, Z.; Li, L.; Bruna, J.
\newblock Supervised Community Detection with Line Graph Neural Networks.
\newblock In Proceedings of the 7th International Conference on Learning
Representations, {ICLR} 2019, New Orleans, LA, USA,  6--9 May 2019.


\bibitem[Liang and Pu(2023)]{Liang2023LineGN}
Liang, J.; Pu, C.
\newblock Line Graph Neural Networks for Link Weight Prediction.
\newblock {\em arXiv} {\bf 2023}, arXiv:2309.15728.

\bibitem[Morshed et~al.(2023)Morshed, Sultana, and Lee]{lelgnn2023}
Morshed, M.G.; Sultana, T.; Lee, Y.K.
\newblock LeL-GNN: Learnable Edge Sampling and Line Based Graph Neural Network
for Link Prediction.
\newblock {\em IEEE Access} {\bf 2023}, {\em 11},~56083--56097. [\href{http://dx.doi.org/10.1109/ACCESS.2023.3283029}{CrossRef}]

\bibitem[Zhang et~al.(2024)Zhang, Xia, Zhang, and Xu]{10004977}
Zhang, H.; Xia, J.; Zhang, G.; Xu, M.
\newblock Learning Graph Representations Through Learning and Propagating Edge
Features.
\newblock {\em IEEE Trans. Neural Netw. Learn. Syst.} {\bf
2024}, {\em 35},~8429--8440. [\href{http://dx.doi.org/10.1109/TNNLS.2022.3228102}{CrossRef}] [\href{http://www.ncbi.nlm.nih.gov/pubmed/37018297}{PubMed}]

\bibitem[Jiang et~al.(2019)Jiang, Ji, and Li]{conf/ijcai/JiangJL19}
Jiang, X.; Ji, P.; Li, S.
\newblock CensNet: Convolution with Edge-Node Switching in Graph Neural
Networks.
\newblock In Proceedings of the IJCAI, {{Macao, 10--16 August 2019}}; Kraus, S., Ed.; {{AAAI Press: Washington, DC, USA}}, 2019; pp.  2656--2662.

\bibitem[Li et~al.(2024)Li, Meng, Ye, Xiao, Cao, and
Zhao]{10.1016/j.jksuci.2024.102011}
Li, M.; Meng, L.; Ye, Z.; Xiao, Y.; Cao, S.; Zhao, H.
\newblock Line graph contrastive learning for node classification.
\newblock {\em J. King Saud Univ. Comput. Inf. Sci.} {\bf 2024}, {\em 36}, {{102011}}. [\href{http://dx.doi.org/10.1016/j.jksuci.2024.102011}{CrossRef}]

\bibitem[Kipf and Welling(2017)]{Kipf:2016tc}
Kipf, T.N.; Welling, M.
\newblock Semi-Supervised Classification with Graph Convolutional Networks.
In Proceedings of the 5th International Conference on Learning
Representations, {ICLR} 2017, Toulon, France,  24--26 April 2017.

\bibitem[Hamilton et~al.(2017)Hamilton, Ying, and Leskovec]{HamiltonYL17}
Hamilton, W.L.; Ying, Z.; Leskovec, J.
\newblock Inductive Representation Learning on Large Graphs.
\newblock In Proceedings of the NIPS, Long Beach, CA, USA, 4--9 December 2017; Guyon, I., von Luxburg, U., Bengio, S.,  Wallach, H.M., Fergus, R., Vishwanathan, S.V.N., Garnett, R., Eds.; {{Curran Associates Inc.: Red Hook, NY, USA}}, 2017;  pp. 1024--1034.

\bibitem[Xu et~al.(2019)Xu, Hu, Leskovec, and
Jegelka]{journals/corr/abs-1810-00826}
Xu, K.; Hu, W.; Leskovec, J.; Jegelka, S.
\newblock How Powerful are Graph Neural Networks?
\newblock In Proceedings of the 7th International Conference on Learning
Representations, {ICLR} 2019, New Orleans, LA, USA,  6--9 May 2019.


\bibitem[Gong et~al.(2023)Gong, Wang, Sun, Liu, Ning, Xiong, and
Peng]{conf/ijcai/GongWSLN0P23}
Gong, Z.; Wang, G.; Sun, Y.; Liu, Q.; Ning, Y.; Xiong, H.; Peng, J.
\newblock Beyond Homophily: Robust Graph Anomaly Detection via Neural
Sparsification.
\newblock In Proceedings of the IJCAI, {{Macao, 19--25 August}} 2023; pp. 2104--2113.

\bibitem[Chien et~al.(2021)Chien, Peng, Li, and
Milenkovic]{conf/iclr/ChienP0M21}
Chien, E.; Peng, J.; Li, P.; Milenkovic, O.
\newblock Adaptive Universal Generalized PageRank Graph Neural Network.
\newblock In Proceedings of the 9th International Conference on Learning
Representations, {ICLR} 2021, Virtual,  Austria,  3--7 May 2021.

\bibitem[Cardoso et~al.(2022)Cardoso, Saleiro, and
Bizarro]{10.1145/3533271.3561727}
Cardoso, M.; Saleiro, P.; Bizarro, P.
\newblock LaundroGraph: Self-Supervised Graph Representation Learning for
Anti-Money Laundering.
\newblock In Proceedings of the Third ACM International
Conference on AI in Finance, ICAIF '22, New York, NY, USA, {{2--4 November}} 2022;  pp. 130--138. [\href{http://dx.doi.org/10.1145/3533271.3561727}{CrossRef}]

\bibitem[Misra et~al.(2016)Misra, Shrivastava, Gupta, and
Hebert]{Misra_2016_CVPR}
Misra, I.; Shrivastava, A.; Gupta, A.; Hebert, M.
\newblock Cross-Stitch Networks for Multi-task Learning.
\newblock In Proceedings of the 2016 {IEEE} Conference on Computer Vision and
Pattern Recognition, {CVPR} 2016, Las Vegas, NV, USA, 27--30 June 2016;
{IEEE} Computer Society: {{Piscataway, NJ, USA}}, 2016; pp. 3994--4003. [\href{http://dx.doi.org/10.1109/CVPR.2016.433}{CrossRef}]

\bibitem[He et~al.(2016)He, Zhang, Ren, and Sun]{He_2016_CVPR}
He, K.; Zhang, X.; Ren, S.; Sun, J.
\newblock Deep Residual Learning for Image Recognition.
\newblock In Proceedings of the 2016 {IEEE} Conference on Computer Vision and
Pattern Recognition, {CVPR} 2016, Las Vegas, NV, USA,  27--30 June 2016;
{IEEE} Computer Society: {{Piscataway, NJ, USA}},  2016; pp. 770--778. [\href{http://dx.doi.org/10.1109/CVPR.2016.90}{CrossRef}]

\bibitem[Kanezashi et~al.(2022)Kanezashi, Suzumura, Liu, and
Hirofuchi]{kanezashi2022ethereum}
Kanezashi, H.; Suzumura, T.; Liu, X.; Hirofuchi, T.
\newblock Ethereum Fraud Detection with Heterogeneous Graph Neural Networks.
\newblock {\em arXiv} {\bf 2022},  arXiv:2203.12363.

\bibitem[Wu et~al.(2022)Wu, Yuan, Lin, You, Chen, Chen, and Zheng]{9184813}
Wu, J.; Yuan, Q.; Lin, D.; You, W.; Chen, W.; Chen, C.; Zheng, Z.
\newblock Who Are the Phishers? Phishing Scam Detection on Ethereum via Network
Embedding.
\newblock {\em IEEE Trans. Syst. Man Cybern. Syst.}
{\bf 2022}, {\em 52},~1156--1166. [\href{http://dx.doi.org/10.1109/TSMC.2020.3016821}{CrossRef}]

\bibitem[Elliott et~al.(2019)Elliott, Cucuringu, Luaces, Reidy, and
Reinert]{Elliott2019AnomalyDI}
Elliott, A.; Cucuringu, M.; Luaces, M.M.; Reidy, P.; Reinert, G.
\newblock Anomaly Detection in Networks with Application to Financial
Transaction Networks.
\newblock {\em arXiv} {\bf 2019}, arXiv:1901.00402.

\bibitem[Corso et~al.(2020)Corso, Cavalleri, Beaini, Li{\`{o}}, and
Velickovic]{CorsoCBLV20}
Corso, G.; Cavalleri, L.; Beaini, D.; Li{\`{o}}, P.; Velickovic, P.
\newblock Principal Neighbourhood Aggregation for Graph Nets.
\newblock In Proceedings of the Advances in Neural Information Processing
Systems 33: Annual Conference on Neural Information Processing Systems 2020,
NeurIPS 2020, Virtual,  6--12 December 2020; Larochelle, H., Ranzato, M.,
Hadsell, R., Balcan, M., Lin, H., Eds.; {{Curran Associates, Inc.: Red Hook, NY, USA, 2020}}.

\bibitem[Kamiński et~al.(2021)Kamiński, Ludwiczak, Jasiński, Bukala, Madaj,
Szczepaniak, and Dunin-Horkawicz]{10.1093/bib/bbab371}
Kamiński, K.; Ludwiczak, J.; Jasiński, M.; Bukala, A.; Madaj, R.;
Szczepaniak, K.; Dunin-Horkawicz, S.
\newblock {Rossmann-toolbox: A deep learning-based protocol for the prediction
and design of cofactor specificity in Rossmann fold proteins}.
\newblock {\em Brief. Bioinform.} {\bf 2021}, {\em 23},~bbab371. [\href{http://dx.doi.org/10.1093/bib/bbab371}{CrossRef}] [\href{http://www.ncbi.nlm.nih.gov/pubmed/34571541}{PubMed}]

\bibitem[Xu et~al.(2018)Xu, Li, Tian, Sonobe, Kawarabayashi, and
Jegelka]{Xu:2018vn}
Xu, K.; Li, C.; Tian, Y.; Sonobe, T.; Kawarabayashi, K.; Jegelka, S.
\newblock Representation Learning on Graphs with Jumping Knowledge Networks.
\newblock In Proceedings of the 35th International
Conference on Machine Learning, {ICML} 2018, {PMLR},
Stockholm, Sweden, 10--15 July 2018; Dy, J.G., Krause, A., Eds.;
PMLR: {{London, UK}}, 2018; Volume~80, pp.  5449--5458.

\bibitem[Hu et~al.(2020)Hu, Liu, Gomes, Zitnik, Liang, Pande, and
Leskovec]{HuLGZLPL20}
Hu, W.; Liu, B.; Gomes, J.; Zitnik, M.; Liang, P.; Pande, V.S.; Leskovec, J.
\newblock Strategies for Pre-training Graph Neural Networks.
\newblock In Proceedings of the 8th International Conference on Learning
Representations, {ICLR} 2020, Addis Ababa, Ethiopia,\linebreak  26--30 April~2020.


\bibitem[He et~al.(2023)He, Zhang, Huang, Rozemberczki, Cucuringu, and
Reinert]{PyTorchGeometricSignedDirected2023}
He, Y.; Zhang, X.; Huang, J.; Rozemberczki, B.; Cucuringu, M.; Reinert, G.
\newblock {{PyTorch Geometric Signed Directed: A Software Package on Graph Neural Networks for Signed and Directed Graphs.}}
\newblock In Proceedings of the Second Learning on Graphs Conference (LoG 2023), {PMLR} 231, Virtual,
New Orleans, LA, USA, 27--30 November~2023.







\bibitem[Paszke et~al.(2019)Paszke, Gross, Massa, Lerer, Bradbury, Chanan,
Killeen, Lin, Gimelshein, Antiga, Desmaison, K{\"{o}}pf, Yang, DeVito,
Raison, Tejani, Chilamkurthy, Steiner, Fang, Bai, and
Chintala]{paszke2019pytorch}
Paszke, A.; Gross, S.; Massa, F.; Lerer, A.; Bradbury, J.; Chanan, G.; Killeen,
T.; Lin, Z.; Gimelshein, N.; Antiga, L.;  et~al.
\newblock PyTorch: An Imperative Style, High-Performance Deep Learning Library.
\newblock In Proceedings of the Advances in Neural Information Processing
Systems 32: Annual Conference on Neural Information Processing Systems 2019,
NeurIPS 2019,  Vancouver, BC, Canada,\linebreak  8--14 December 2019; Wallach, H.M.,
Larochelle, H., Beygelzimer, A., d'Alch{\'{e}}{-}Buc, F., Fox, E.B., Garnett,
R., Eds.; {{Curran Associates, Inc.: Red Hook, NY, USA, 2019}}; pp. 8024--8035.

\bibitem[Wang et~al.(2019)Wang, Zheng, Ye, Gan, Li, Song, Zhou, Ma, Yu, Gai,
Xiao, He, Karypis, Li, and Zhang]{wang2020deep}
Wang, M.; Zheng, D.; Ye, Z.; Gan, Q.; Li, M.; Song, X.; Zhou, J.; Ma, C.; Yu,
L.; Gai, Y.;  et~al.
\newblock Deep Graph Library: A Graph-Centric, Highly-Performant Package for
Graph Neural Networks.
\newblock {\em arXiv} {\bf 2019}, arXiv:1909.01315.

\bibitem[Fey and Lenssen(2019)]{fey2019graph}
Fey, M.; Lenssen, J.E.
\newblock Fast Graph Representation Learning with PyTorch Geometric.
\newblock {\em arXiv} {\bf 2019},  arXiv:1903.02428.


\bibitem[Kingma and Ba(2015)]{KingBa15}
Kingma, D.P.; Ba, J.
\newblock Adam: {A} Method for Stochastic Optimization.
In Proceedings of the 3rd International Conference on Learning
Representations, {ICLR} 2015, San Diego, CA, USA,  7--9 May 2015.
% ; Bengio, Y., LeCun, Y., Eds.;  {2015}


\bibitem[Loshchilov and Hutter(2017)]{conf/iclr/LoshchilovH17}
Loshchilov, I.; Hutter, F.
\newblock {SGDR:} Stochastic Gradient Descent with Warm Restarts.
In Proceedings of the 5th International Conference on Learning
Representations, {ICLR} 2017, Toulon, France,  24--26 April 2017.

\end{thebibliography}
\end{document}